\documentclass[sigconf]{acmart}

\usepackage{subcaption}
\usepackage{caption}
\usepackage{multirow}

\newcommand{\ve}[1]{\mathbf{#1}}

\usepackage{bm}

\AtBeginDocument{%
  \providecommand\BibTeX{{%
    \normalfont B\kern-0.5em{\scshape i\kern-0.25em b}\kern-0.8em\TeX}}}

\setcopyright{none}
\renewcommand\footnotetextcopyrightpermission[1]{}  
\copyrightyear{2021}
\acmYear{2021}

\begin{document}

\title{Discriminative Distillation to Reduce Class Confusion in Continual Learning}




\author{Changhong~Zhong, Zhiying~Cui, Ruixuan~Wang, and~Wei-Shi Zheng}
\thanks{C. Zhong, Z. Cui, R. Wang, and WS. Zheng are with the School of Computer Science and Engineering, Sun Yat-sen University, and also with the Key Laboratory of Machine Intelligence and Advanced Computing, MOE, Guangzhou, 510000, China; \\
Correspondence e-mail: wangruix5@mail.sysu.edu.cn}

\renewcommand{\shortauthors}{C. Zhong and Z. Cui, et al.}

\begin{abstract}

Successful continual learning of new knowledge would enable intelligent systems to recognize more and more classes of objects. However, current intelligent systems often fail to correctly recognize previously learned classes of objects when updated to learn new classes. It is widely believed that such downgraded performance is solely due to the catastrophic forgetting of previously learned knowledge. In this study, we argue that the class confusion phenomena may also play a role in downgrading the classification performance during continual learning, i.e., the high similarity between new classes and any previously learned classes would also cause the classifier to make mistakes in recognizing these old classes, even if the knowledge of these old classes is not forgotten. To alleviate the class confusion issue, we propose a discriminative distillation strategy to help the classify well learn the discriminative features between confusing classes during continual learning. Experiments on multiple natural image classification tasks support that the proposed distillation strategy, when combined with existing methods, is effective in further improving continual learning.

\end{abstract}




\maketitle
\pagestyle{plain}

\section{Introduction}

\begin{figure}[t]
\begin{center}
  \includegraphics[width=0.45\linewidth]{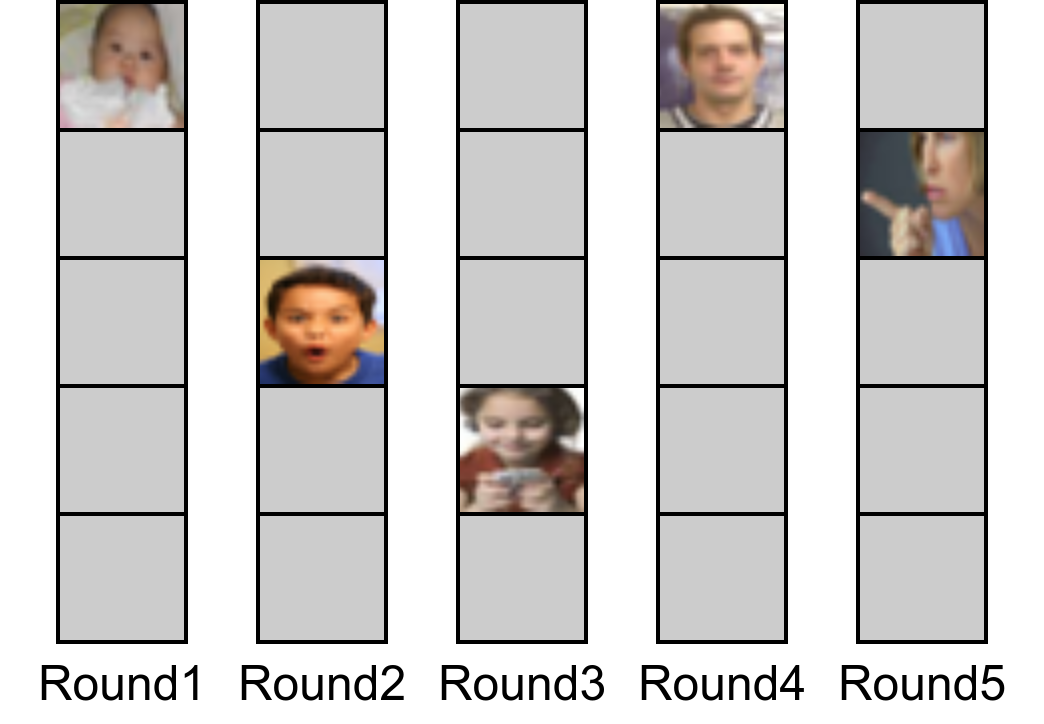}
  \includegraphics[width=0.50\linewidth]{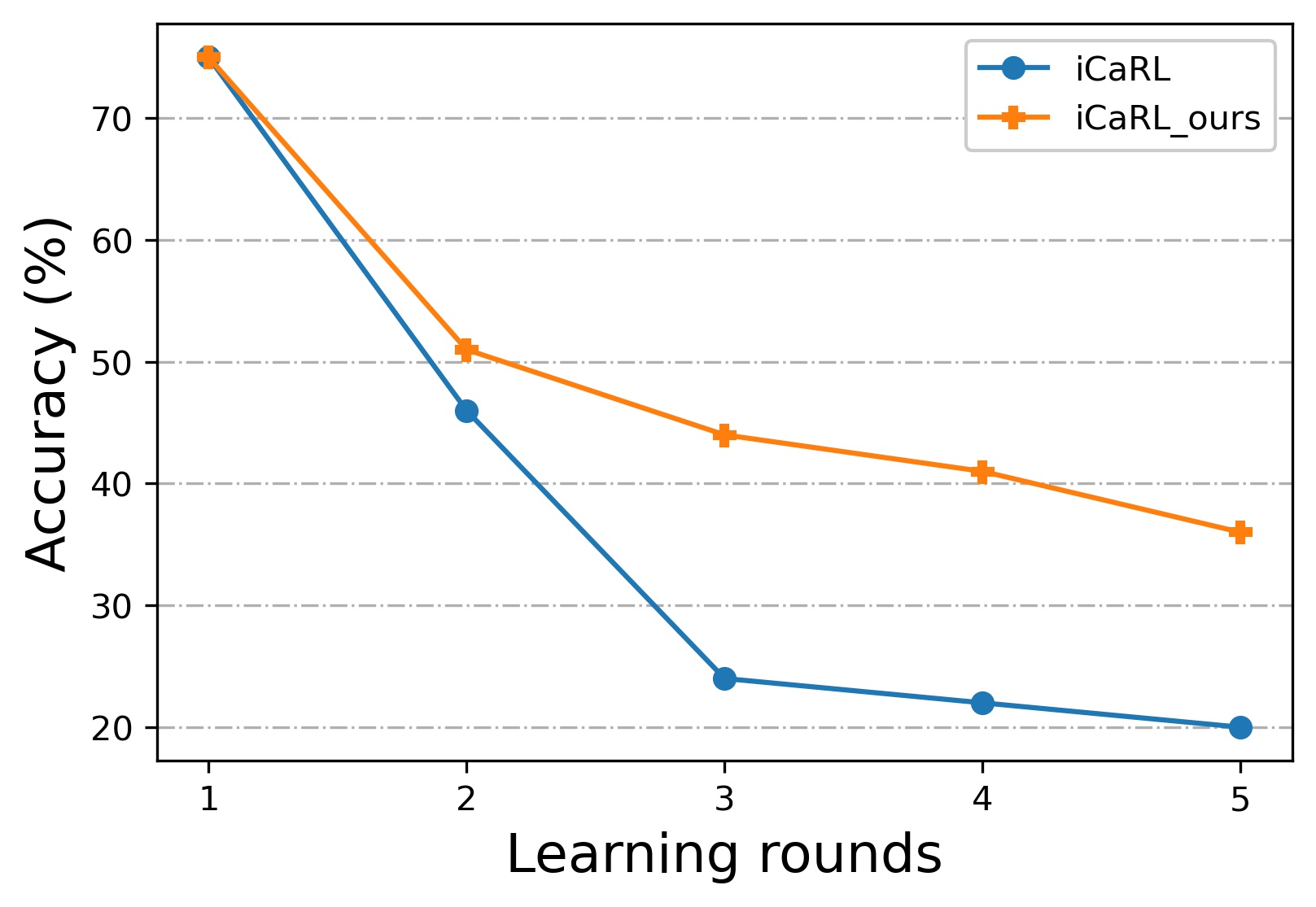}
\end{center}
  \caption{Continual learning over multiple rounds suffers not only from catastrophic forgetting but also from the confusion between old and new classes. Left: five similar classes (`baby', `boy', `girl', `man', and `woman') were learned at different rounds of continual learning; gray boxes represent certain other classes learned at each round. Right: the classification performance on the `baby' class learned at the first round decreases over learning rounds, but the proposed method (orange curve) can better handle the confusion between the `baby' class and its similar classes at later rounds compared to the baseline iCaRL (blue curve).}
\label{fig:motivation}
\end{figure}

Continual learning or lifelong learning aims to continually learn and absorb new knowledge over time while retaining previously learned knowledge~\cite{parisi2019continual}. With this ability, humans can accumulate knowledge over time and become experts in certain domains. It is desirable for the intelligent system to obtain this ability and recognize more and more objects continually, with the presumption that very limited amount or even no data is stored for the old classes when learning knowledge of new classes. 
The intelligent system has to update its parameters when acquiring new knowledge and often inevitably causes the downgraded performance on recognizing old classes. It has been widely believed that the downgraded performance is solely due to the \textit{catastrophic forgetting} of old knowledge during learning new knowledge~\cite{kemker2018measuring,kirkpatrick2017overcoming}, and various approaches have been proposed to alleviate the catastrophic forgetting issue, such as by trying to keep important model parameters or outputs at various layers in convolutional neural networks (CNNs) unchanged during learning new knowledge~\cite{kirkpatrick2017overcoming,li2017learning,dhar2019learning,douillard2020podnet,hou2019learning}.

However, sometimes simply keeping old knowledge from forgetting during continual learning may not be enough to keep classification performance from downgrading. At an early round of continual learning, since only a few classes of knowledge needs to be learned, the classifier may easily learn to use part of class knowledge to well discriminate between these classes. 
When any new class is visually similar to any previously learned class during continual learning, the visual features learned to recognize the old class may not be discriminative enough to discriminate between the new class and the visually similar old class (e.g., `girl' vs. `baby' face images, Figure~\ref{fig:motivation}), causing the downgraded performance on the previously learned class. We call this phenomena the \textit{class confusion issue}. In this study,  we propose a novel knowledge distillation strategy to help the classifier learn such discriminative knowledge information between old and new classes during continual learning. The basic idea is to train a temporary expert classifier to learn both the new classes and visually similar old classes during continual learning, and then distill the discriminative knowledge from the temporary expert classifier to the new classifier. To the best of our knowledge, it is the first time to explore the class confusion issue in continual learning.
The main contributions of this study are summarized below:
\begin{itemize}
    \item It is observed that continual learning is affected not only by catastrophic forgetting, but also by  potential class confusion between new classes and visually similar old classes.
    \item A discriminative knowledge distillation strategy is proposed to help the classifier discriminate confusing classes. 
    \item Initial experiments on multiple image classification datasets support that the proposed discriminative distillation can be flexibly combined with existing methods and is effective in improving continual learning.
\end{itemize}





\section{Related work}

Generally, there are two types of continual learning problems, task-incremental and class-incremental. Task-incremental learning presumes that one model is incrementally updated to solve more and more tasks, often with multiple tasks sharing a common feature extractor but having task-specific classification heads. The task identification is available during inference, i.e., users know which model head should be applied when predicting the class label of a new data. In task-incremental learning, confusion issue is naturally solved due to the acknowledgement of certain task identification and separated classification heads. In contrast, class-incremental learning presumes that one model is incrementally updated to predict more and more classes, with all classes sharing a single model head. Therefore, the confusion issue potentially affect the continual learning procedure. This study focuses on the class-incremental learning problem. Existing approaches to the two types of continual learning can be roughly divided into four groups, regularization-based, expansion-based, distillation-based, and regeneration-based.

Regularization-based approaches often find the model parameters or components (e.g., kernels in CNNs) crucial for old knowledge, and then try to keep them unchanged with the help of regularization loss terms when learning new classes~\cite{abati2020conditional,ahn2019uncertainty,fernando2017pathnet,jung2020continual,kim2018keep,kirkpatrick2017overcoming,mallya2018packnet}. The importance of each model parameter can be measured by the sensitivity of the loss function to changes in model parameters as in the elastic weight consolidation (EWC) method~\cite{kirkpatrick2017overcoming}, or by the sensitivity of the model output to small changes in model parameter as in the memory aware synapses (MAS) method~\cite{aljundi2018memory}. The importance of each kernel in a CNN model can be measured based on the magnitude of the kernel (e.g., L2 norm of the kernel matrix) as in the PackNet~\cite{mallya2018packnet}.
While keeping the parameters unchanged could help models keep old knowledge in few rounds of continual learning, it is not able to solve the confusion issue because more and more parameters in CNNs are frozen. The frozen parameters keep the extracted features unchanged and not able to provide discriminative enough information to distinguish between similar classes.

To make models more flexibly learn new knowledge, expansion-based approaches are developed by adding new kernels, layers, or even sub-networks when learning new knowledge~\cite{aljundi2017expert,hung2019compacting,karani2018lifelong,li2019learn,rajasegaran2019random,yoon2017lifelong}. 
For example, Aljundi et al.~\cite{aljundi2017expert} proposed expanding an additional network for a new task and training an expert model to make decisions about which network to use during inference. It turns a class-incremental learning problem to a task-incremental problem at the cost of additional parameters. As another example, Yoon et al. proposed a dynamically expandable network (DEN)~\cite{yoon2017lifelong} by selectively  retraining the network and expanding kernels at each layer if necessary. Although expanding the network architecture can potentially alleviate the confusion issue to some extent because the expanded kernels might help extract more discriminative features, most expansion-based approaches are initially proposed for task-incremental learning and might not flexible to be extended for class-incremental learning.

In comparison, distillation-based approaches can be directly applied to class-incremental learning by distilling knowledge from the old classifier (for old classes) to the new classifier (for both old and new classes) during learning new classes~\cite{castro2018end,hou2018lifelong,iscen2020memory,li2017learning,meng2020adinet,rebuffi2017icarl}, where the old knowledge is implicitly represented by soft outputs of the old classifier with stored small number of old images and/or new classes of images as the inputs. A distillation loss is added to the original cross-entropy loss during learning the new classifier, where the distillation loss helps the new classifier have similar relevant output compared to the output of the old classifier for any input image. The well-known methods include the learning without forgetting (LwF)~\cite{li2017learning}, the incremental classifier and representation learning (iCaRL)~\cite{rebuffi2017icarl}, and the end-to-end incremental learning (End2End)~\cite{castro2018end}. More recently, the distillation has been extended to intermediate CNN layers, either by keeping feature map activations unchanged as in the learning without memorizing (LwM)~\cite{dhar2019learning}, or by keeping the spatial pooling unchanged respectively along the width and the height directions as in PODNet~\cite{douillard2020podnet}, or by keeping the normalized global pooling unchanged at last convolutional layer as in the unified classifier incrementally via rebalancing (UCIR)~\cite{hou2019learning}. These distillation-based methods achieve the state-of-the-art performance for the class-incremental learning problem.

In addition, regeneration-based approaches have also been proposed particularly when none of old-class data is available during learning new classes. The basic idea is to train an auto-encoder~\cite{hayes2019remind,rao2019continual,riemer2019scalable} or generative adversarial network (GAN)~\cite{ostapenko2019learning,rios2018closed,shin2017continual,xiang2019incremental} to synthesize old data for each old class, such that plenty of synthetic but realistic data for each confusion old class can provide detailed information for distinguishing between confusion classes. However, such methods heavily depends on the quality of synthetic data, and data quality could become worse especially when using a single regeneration model for more and more old classes. All the existing approaches are proposed to alleviate the catastrophic forgetting issue, without aware of the existence of the class confusion issue.



\section{Method}
In contrast to most continual learning methods which only aim to reduce catastrophic forgetting during learning new classes, this study additionally aims to reduce the potential confusions between new classes and visually similar old classes. As in most distillation-based continual learning methods, only a small subset of training data is stored for each old class and available during continual learning of new classes.





\subsection{Overview of the proposed framework}
\begin{figure*}[!tbh]
	\centering
	\includegraphics[width=0.9\linewidth, ]{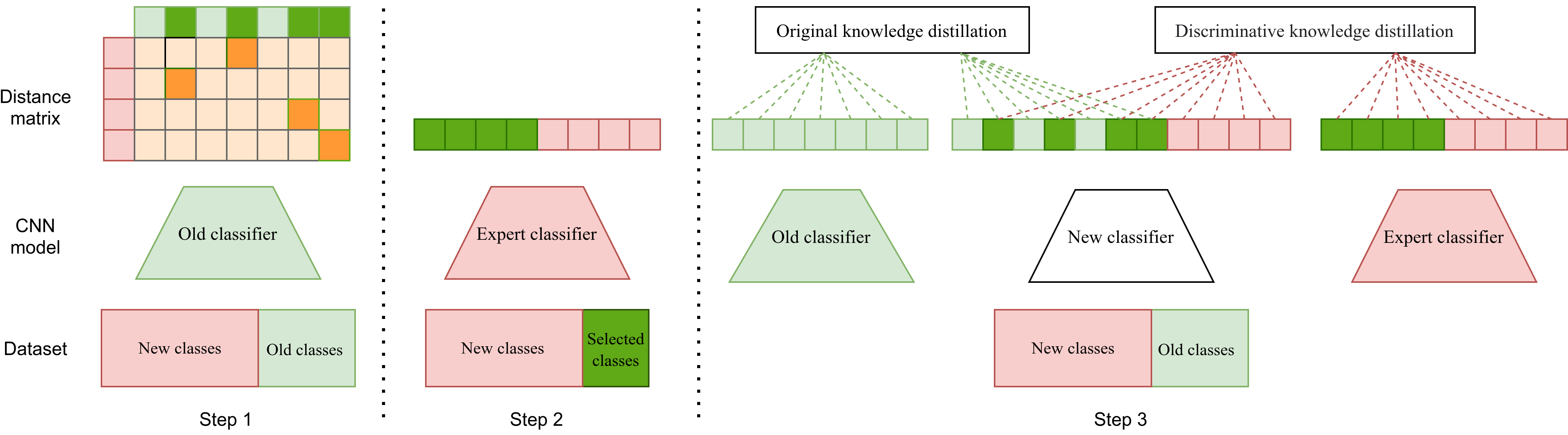}    
	\caption{Discriminative knowledge distillation pipeline. First, the old classifier learned at the previous round is used to identify the similar old class(es) for each new class (Step 1). Then, the temporary expert classifier is trained to recognize both the new classes and their similar old classes (Step 2). Finally, the old classifier and the expert classifier are simultaneously used to teach the new classifier (Step 3). The potential confusion between new and old classes can be alleviated by the distillation from the expert classifier to the new classifier.}
	\label{fig:pipeline}
\end{figure*}

We propose a distillation strategy particularly to reduce the class confusion issue during continual learning. At each round of continual learning, besides the knowledge distillation from the old classifier learned at the previous round to the new classifier at the current round, a temporary expert classifier is trained to classify not only the new classes but also those old classes which are visually similar to the new classes (Figure~\ref{fig:pipeline}, Step 2), and then the discriminative knowledge of the expert classifier is distilled to the new classifier as well during training the new classifier (Figure~\ref{fig:pipeline}, Step 3). 
The knowledge distillation from the expert classifier to the new classifier would largely reduce the potential confusion between these similar classes during prediction by the new classifier. 
It is worth noting that the discriminative knowledge distillation from the expert can be used as a plug-in component for most distillation-based continual learning methods. 



\subsection{Expert classifier}

The key novelty of the proposed framework is the addition of the expert classifier whose knowledge will be distilled to the new classifier. The expert classifier at each learning round is trained to classify both the new classes at the current round and those old classes which are  similar to and therefore more likely confused with the new classes. In this way, the discriminative knowledge between such similar classes can be explicitly learned, and the distillation of such discriminative knowledge would likely reduce the confusion between each new class and its similar old class(es). 

To find the old class(es) similar to each new class, the feature extractor part of the old classifier is used to output the feature representation of each new-class data and stored old-class data, and then the class-centre representation is  obtained respectively for each class by averaging the feature representations of all data belonging to the same class. The Euclidean distance from the class-centre representation of the new class to that of each old class is then used to select the most similar (i.e., closest) old class(es) for the new class (Figure~\ref{fig:pipeline}, Step 1).
While sometimes one new class may have multiple similar old classes and another new class may have no similar old classes, without loss of generality, the same number of similar old classes is selected for each new class in this study and no old class is selected multiple times at each learning round.

Once the old classes similar to the new classes are selected, the expert classifier can be trained using all the training data of the new classes and the stored similar old-class data (Figure~\ref{fig:pipeline}, Step 2). Since only very limited number of old data is available for each old class, the training data set is imbalanced across classes, which could make the classifier focus on learning knowledge of the large (i.e., new) classes. To alleviate the imbalance issue, the expert classifier is initially trained (for 80 epochs in this study) using all the available training set and then fine-tuned (for 40 epochs in in this study) with balanced dataset across classes by down-sampling the dataset of new classes.



\subsection{Knowledge distillation}

The expert classifier, together with the old classifier from the previous round of continual learning, is used to jointly teach the new classifier based on the knowledge distillation strategy.
Suppose $D=\{(\ve{x}_i, \ve{y}_i), i=1,\ldots,N\}$ is the collection of all new classes of training data at current learning round and the stored small old-class data, where $\ve{x}_i$ is an image and the one-hot vector $\ve{y}_i$ is the corresponding class label. For image $\ve{x}_i$, let $\ve{z}_i = [z_{i1},z_{i2},\ldots, z_{it}]^\mathsf{T}$ denote the logit output (i.e., the input to the last-layer softmax operation in the CNN classifier) of the expert classifier, and $\hat{\ve{z}}_i = [\hat{z}_{i1},\hat{z}_{i2},\ldots, \hat{z}_{it}]^\mathsf{T}$ denote the corresponding logit output of the new classifier (Figure~\ref{fig:pipeline}, Step 3, outputs of the new classifier with dashed red lines linked),
where $t$ is the number of outputs by the expert classifier. 
Then, the distillation of the knowledge from the expert classifier to the new classifier can be obtained by minimizing the distillation loss $\mathcal{L}_n$,
\begin{equation}
\mathcal{L}_{n}(\bm{\theta})= -\frac{1}{N} \sum_{i=1}^N \sum_{j=1}^t p_{ij} \log \hat{p}_{ij} \,,
\end{equation}
where $\bm{\theta}$ represents the model parameters of the new classifier, and $p_{ij}$ and $\hat{p}_{ij}$ are from the temperature-tuned softmax operation,
\begin{equation}
p_{ij}= \frac{\exp{(z_{ij}/T_n)}}{\sum_{k=1}^t \exp{(z_{ik}/T_n)}} \,, \quad
\hat{p}_{ij}= \frac{\exp{(\hat{z}_{ij}/T_n)}}{\sum_{k=1}^t \exp{(\hat{z}_{ik}/T_n)}} \,,
\end{equation}
and $T_n\ge 1$ is the temperature coefficient used to help knowledge distillation~\cite{hinton2015distilling}. Since the expert has been trained to discriminate new classes from visually similar old classes, the knowledge distillation from the expert classifier to the new classifier is expected to help the new classifier gain similar discriminative power. In other words, with the distillation, the new classifier would become less confused with the new classes and  visually similar old classes, resulting in better classification performance after each round of continual learning.

Besides the knowledge distillation from the expert classifier, knowledge from the old classifier can be distilled to the new classifier in a similar way, i.e., by minimizing the distillation loss $\mathcal{L}_o$,
\begin{equation}
\mathcal{L}_{o}(\bm{\theta})= -\frac{1}{N} \sum_{i=1}^N \sum_{j=1}^{s} q_{ij} \log \hat{q}_{ij} \,,
\end{equation}
where $s$ is the number of old classes learned so far, and $q_{ij}$ and $\hat{q}_{ij}$ are respectively from the temperature-tuned softmax over the logit of the old classifier and the corresponding logit part of the new classifier (Figure~\ref{fig:pipeline}, Step 3, outputs of the new classifier with dashed green lines linked),
with the distillation parameter $T_o$.

As in general knowledge distillation strategy, besides the two distillation losses, the cross-entropy loss $\mathcal{L}_c$ over the training set $D$ based on the output of the new classifier  is also applied to train the new classifier.
In combination, the new classifier can be trained by minimizing the loss $\mathcal{L}$,
\begin{equation}
\mathcal{L}(\bm{\theta})= \mathcal{L}_{c}(\bm{\theta}) + \lambda_{1} \mathcal{L}_{o}(\bm{\theta}) + \lambda_{2} \mathcal{L}_{n}(\bm{\theta}) \,,
\end{equation}
where $\lambda_{1}$ and $\lambda_{2}$ are trade-off coefficients to balance the loss terms.

The proposed distillation strategy is clearly from existing distillations for continual learning. Most distillation-based continual learning methods only distill knowledge from the old class to the new class at each learning round. The most relevant work is the dual distillation~\cite{li2020continual}, where the expert classifier is trained only for new classes and then, together with the old classifier, distilled to the new classifier. In comparison, the expert classifier in our method is trained to learn not only the new classes but also likely confusing old classes. Therefore, our method can be considered as an extension of the dual distillation strategy, particularly aiming to alleviate the class confusion issue. Mostly importantly, the proposed discriminative distillation can be easily combined with most existing continual learning methods by simply adding the loss term $\mathcal{L}_{n}(\bm{\theta})$ during classifier training at each round of continual learning.


\section{Experiments}

\subsection{Experimental settings}
The proposed method was evaluated on three datasets, CIFAR100~\cite{krizhevsky2009learning}, the full ImageNet dataset~\cite{deng2009imagenet}, and a subset of ImageNet which contains randomly selected 100 classes (Table~\ref{tab:dataset}). 
During model training, each CIFAR100 image was randomly flipped horizontally, and each ImageNet image was randomly cropped and then resized to $224 \times 224$ pixels.
On each dataset, an CNN classifier was first trained for certain number (e.g., 10, 20) of classes, and then a set of new classes' data were provided to update the classifier at each round of continual learning. 
Ths SGD optimizer (batch size 128) was used with an initial learning rate 0.1. The new classifier at each round of continual learning was trained for up to 100 epochs, with the training convergence consistently observed. 
ResNet32 and ResNet18 were used as the default CNN backbone for CIFAR100 and ImageNet (including mini-ImageNet) respectively, and $\lambda_1=\lambda_2=1.0$, $T_n=T_o=2.0$.
One similar old class was selected for each new class in the expert classifier. Following  iCaRL~\cite{rebuffi2017icarl}, the herding strategy was adopted to select a small subset of images for each new class with a total memory size $K$. For CIFAR100 and mini-ImageNet, the memory size is $K=2000$. And for ImageNet, the memory size is $K=20000$. 

\begin{table}[tbh]
    \centering
    \caption{Statistics of datasets used in experiments. $[75, 2400]$ is the size range of image height and width.}
    \begin{tabular}{ccccc}
        \toprule
         Dataset   & \#class & Train/class & Test/class &  Size \\ 
         \midrule
         CIFAR100  & 100   & 500    & 100  & 32$\times$32  \\
         mini-ImageNet &100 & $\sim$1,200  & 100  & [75, 2400]  \\
         ImageNet  & 1000   & $\sim$1,200 & 100 & [75, 2400]\\
         \bottomrule
    \end{tabular}
    \label{tab:dataset}
\end{table}


After training at each round, the average accuracy over all learned classes so far was calculated. Such a training and evaluation process was repeated in next-round continual learning. For each experiment, the average accuracy over three runs were reported, each run with a different and fixed order of classes to be learned. All baseline methods were evaluated on the same orders of continual learning over three runs and with the same herding strategy for testing.

\subsection{Effectiveness evaluation}

\begin{figure}[h]
    \centering
    \includegraphics[width=0.24\linewidth]{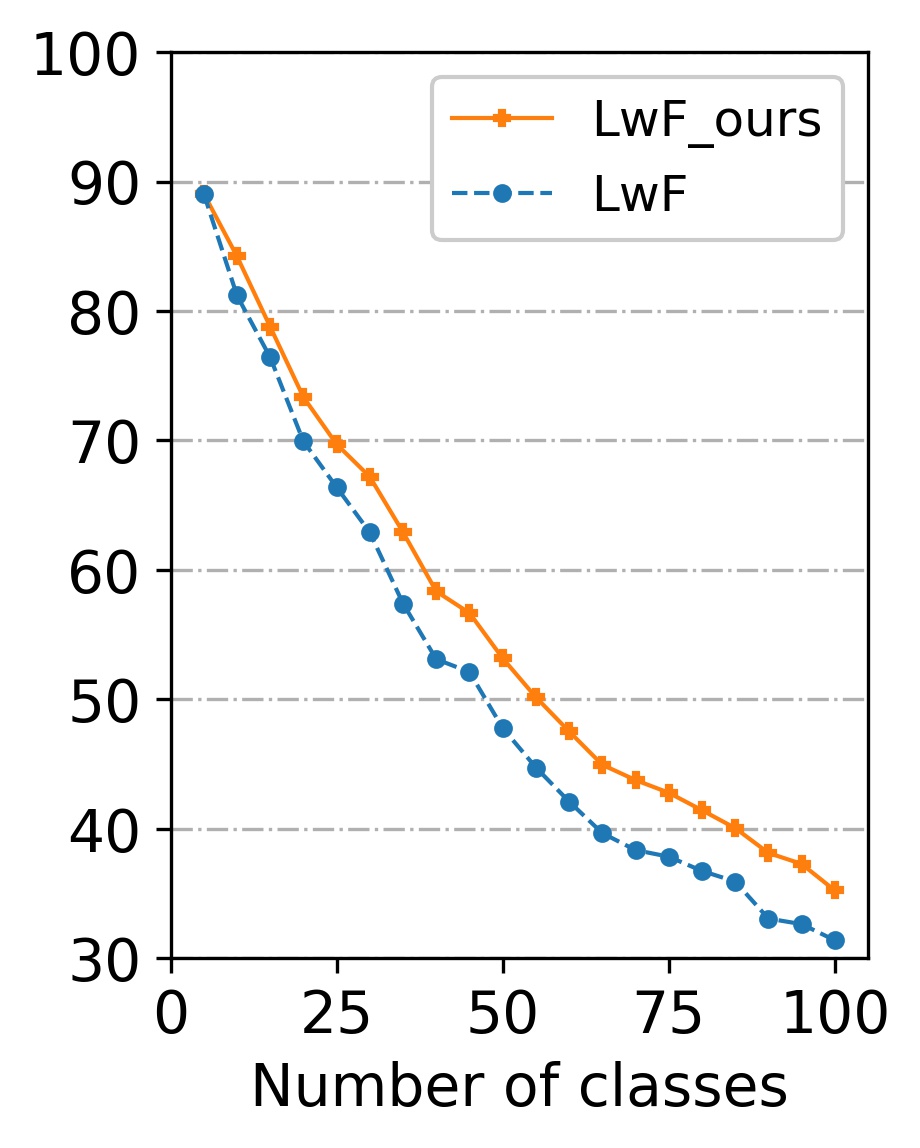}
    \includegraphics[width=0.24\linewidth]{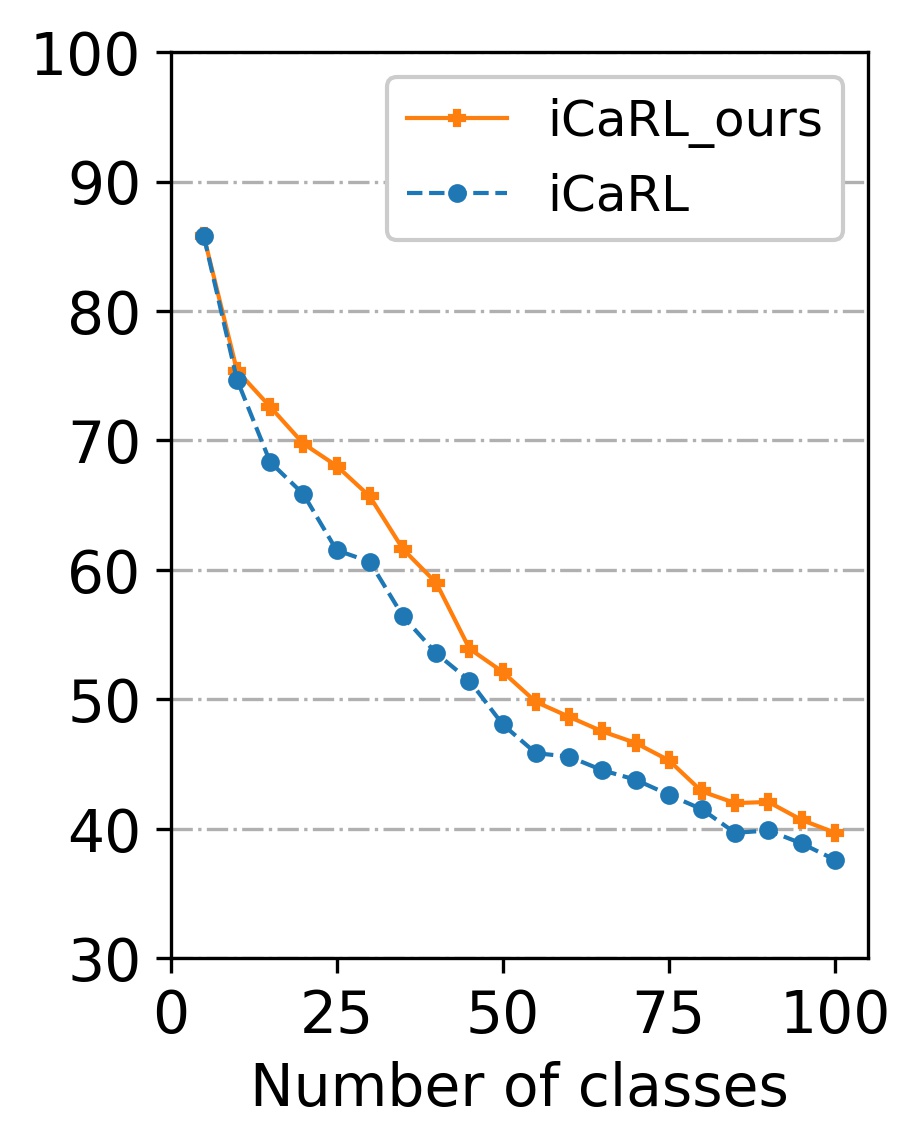}
    \includegraphics[width=0.24\linewidth]{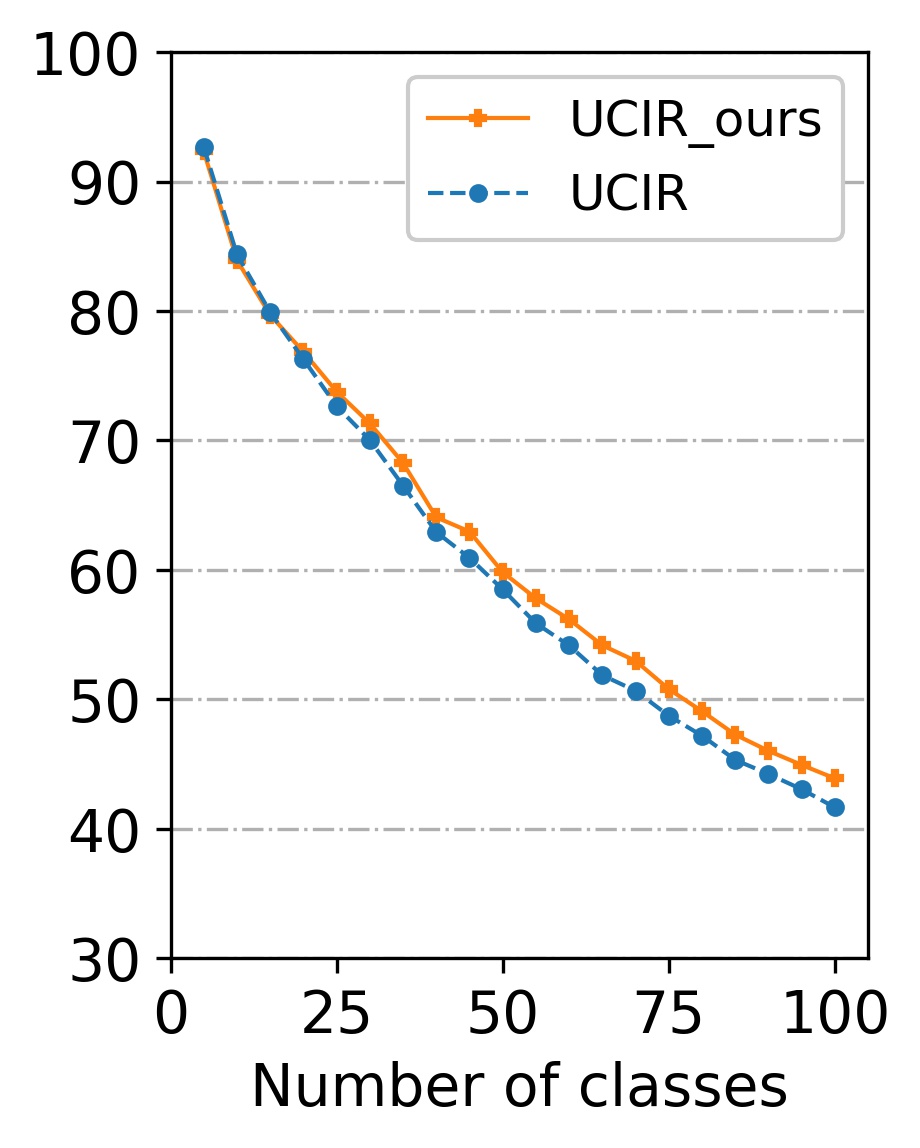}
    \includegraphics[width=0.24\linewidth]{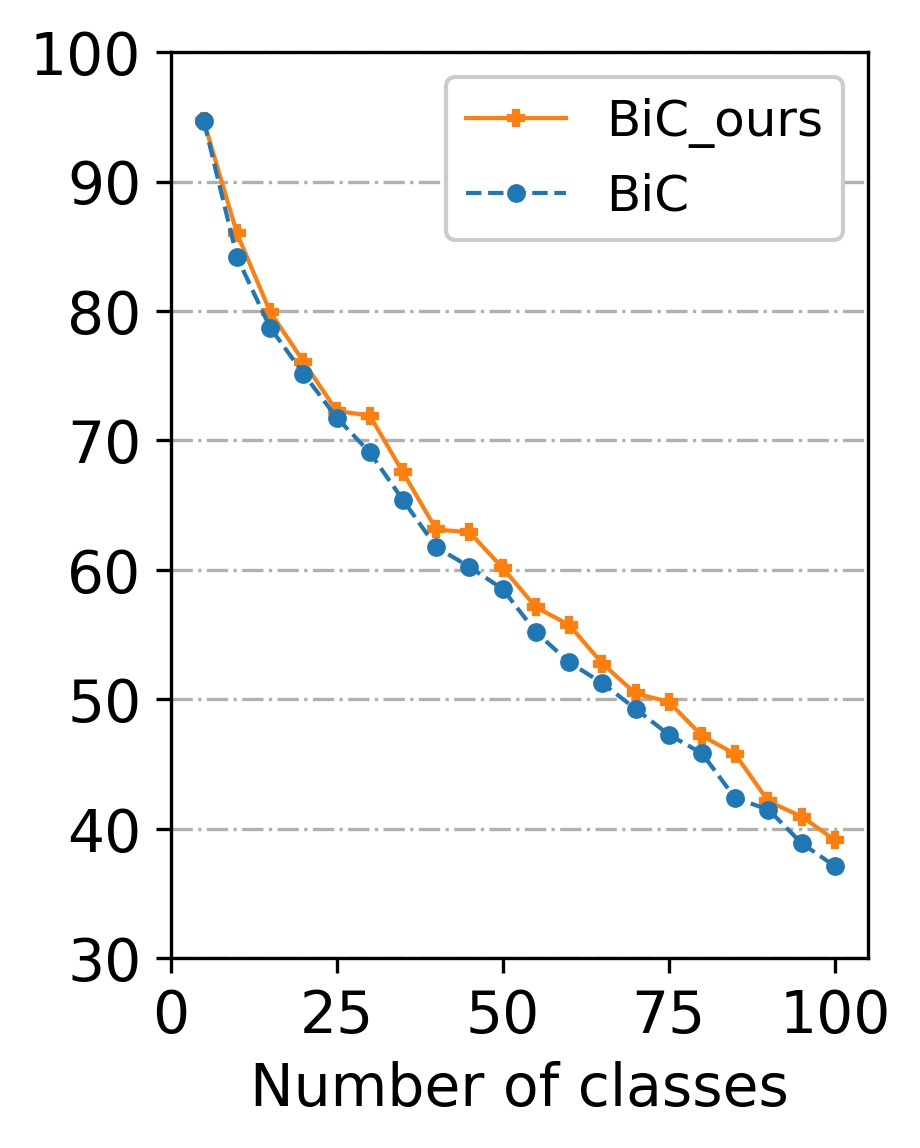}
    \includegraphics[width=0.24\linewidth]{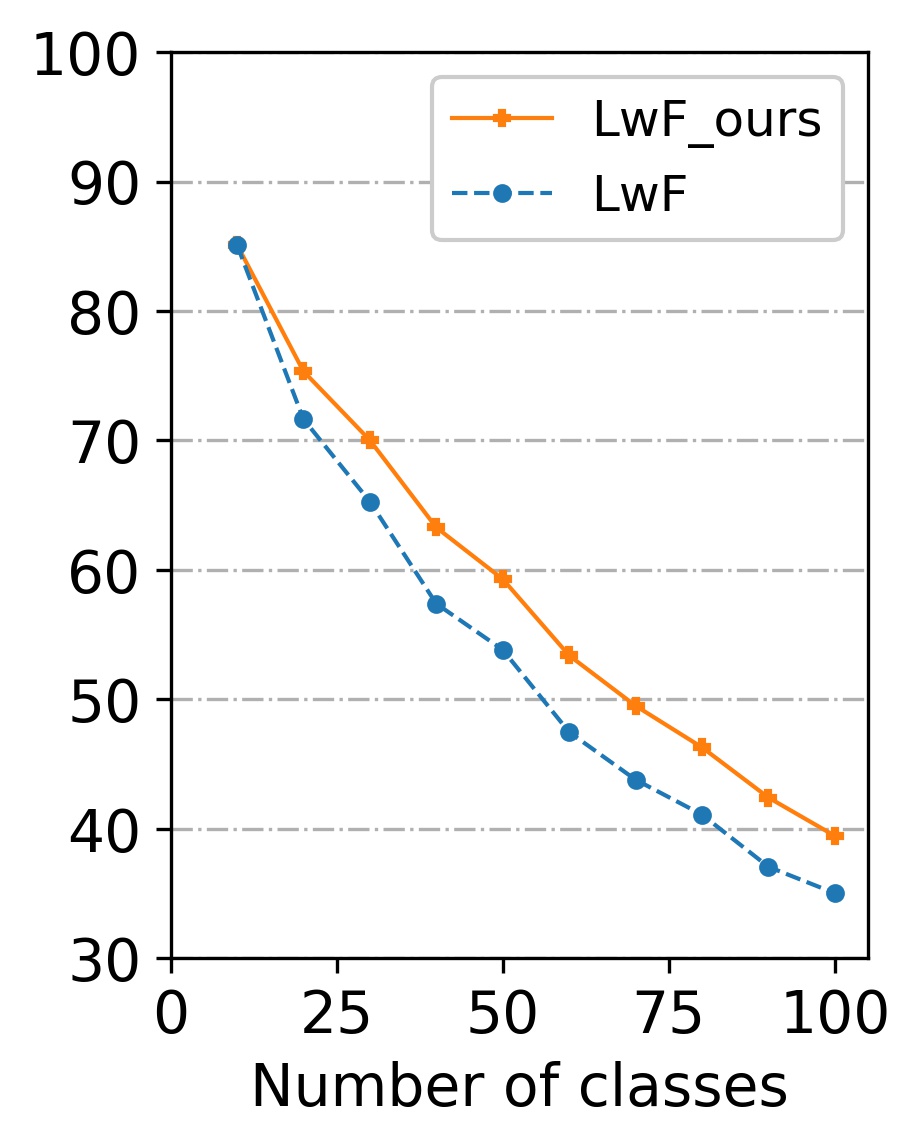}
    \includegraphics[width=0.24\linewidth]{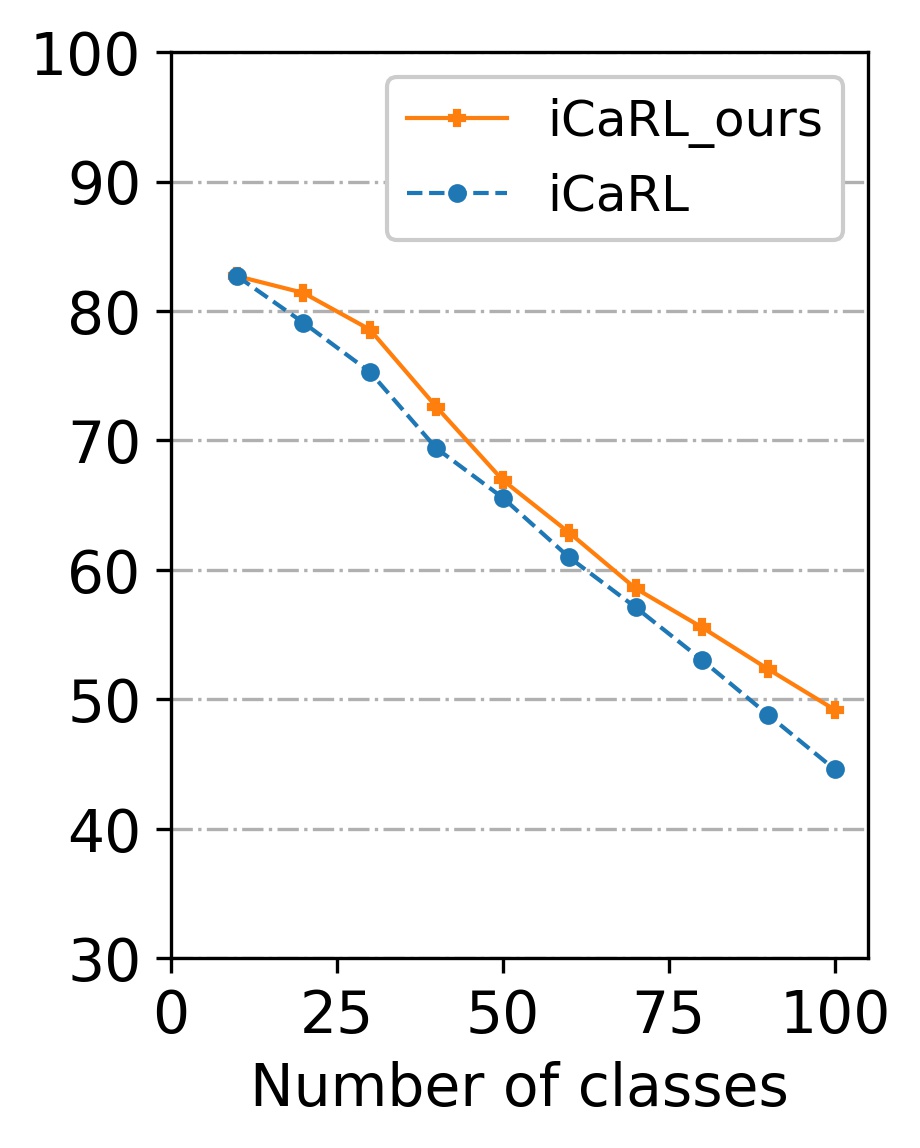}
    \includegraphics[width=0.24\linewidth]{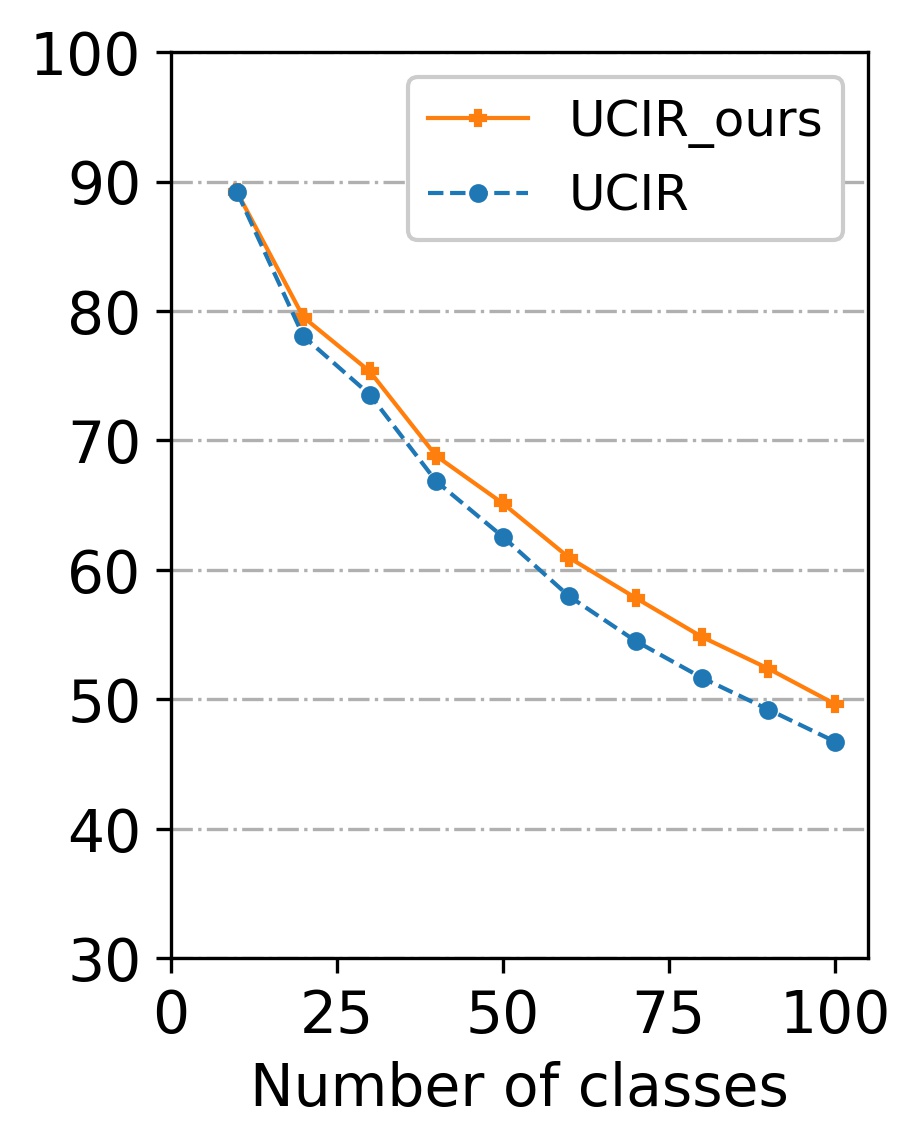}
    \includegraphics[width=0.24\linewidth]{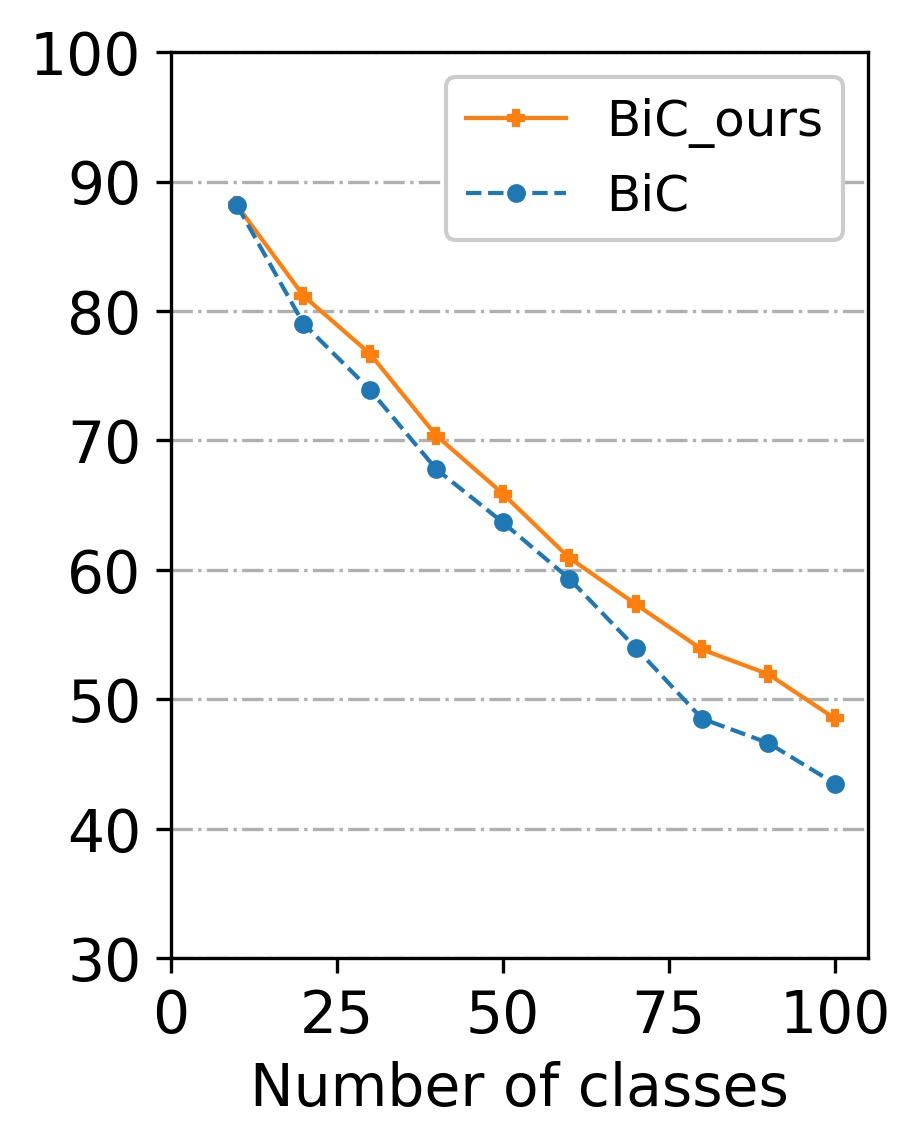}
    \includegraphics[width=0.24\linewidth]{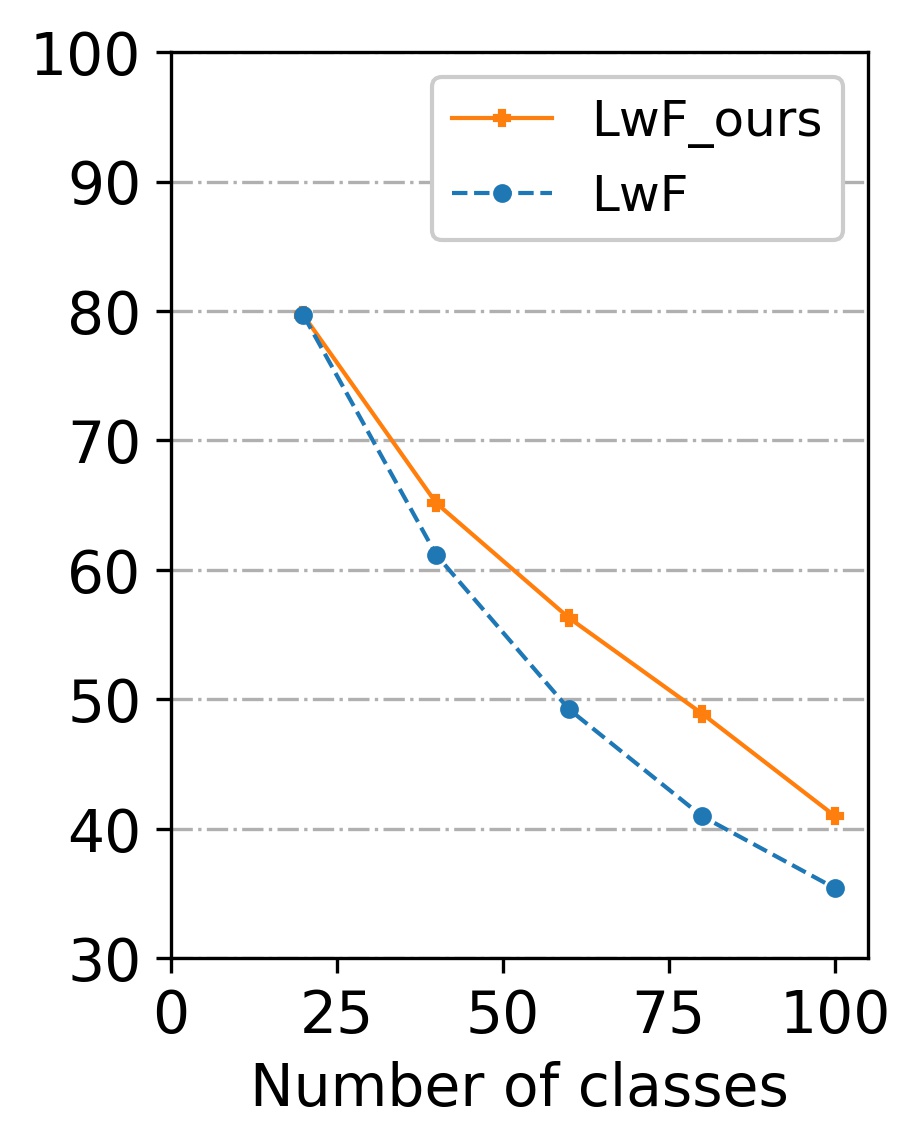}
    \includegraphics[width=0.24\linewidth]{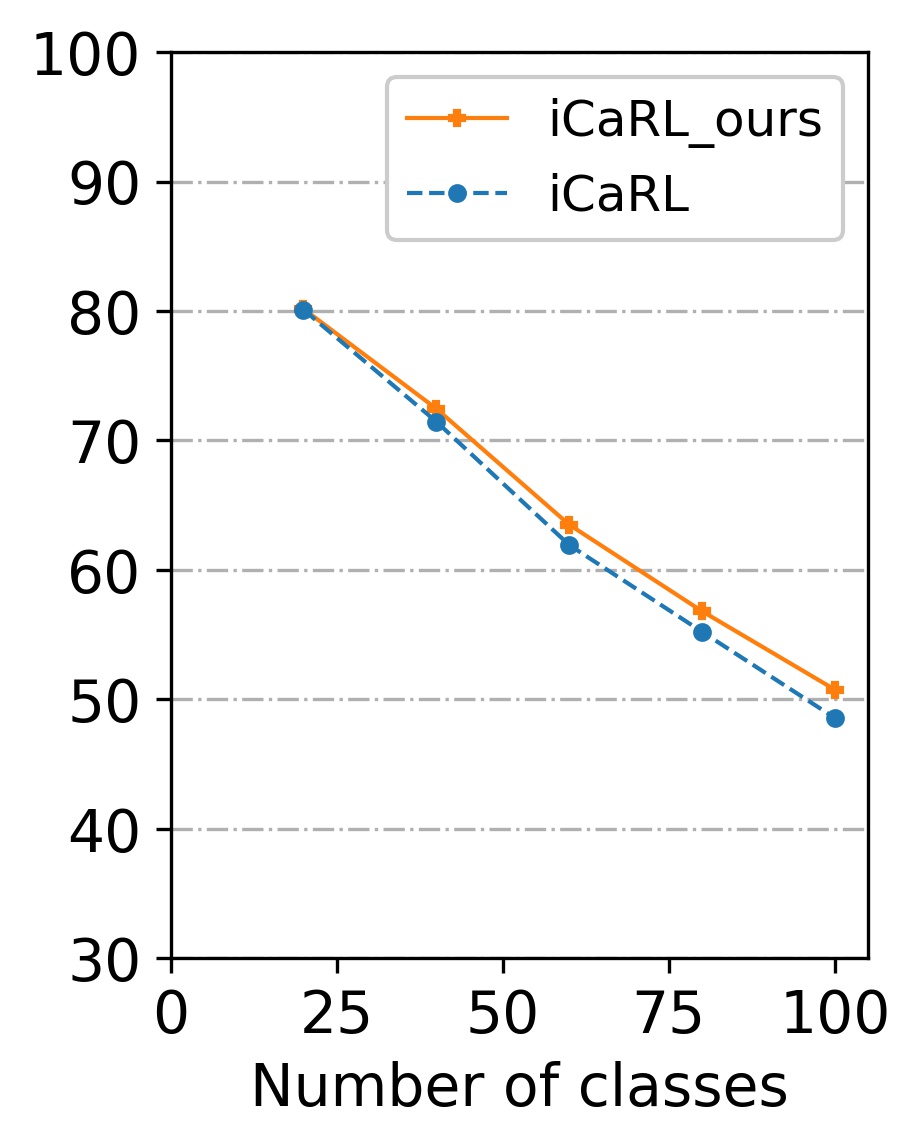}
    \includegraphics[width=0.24\linewidth]{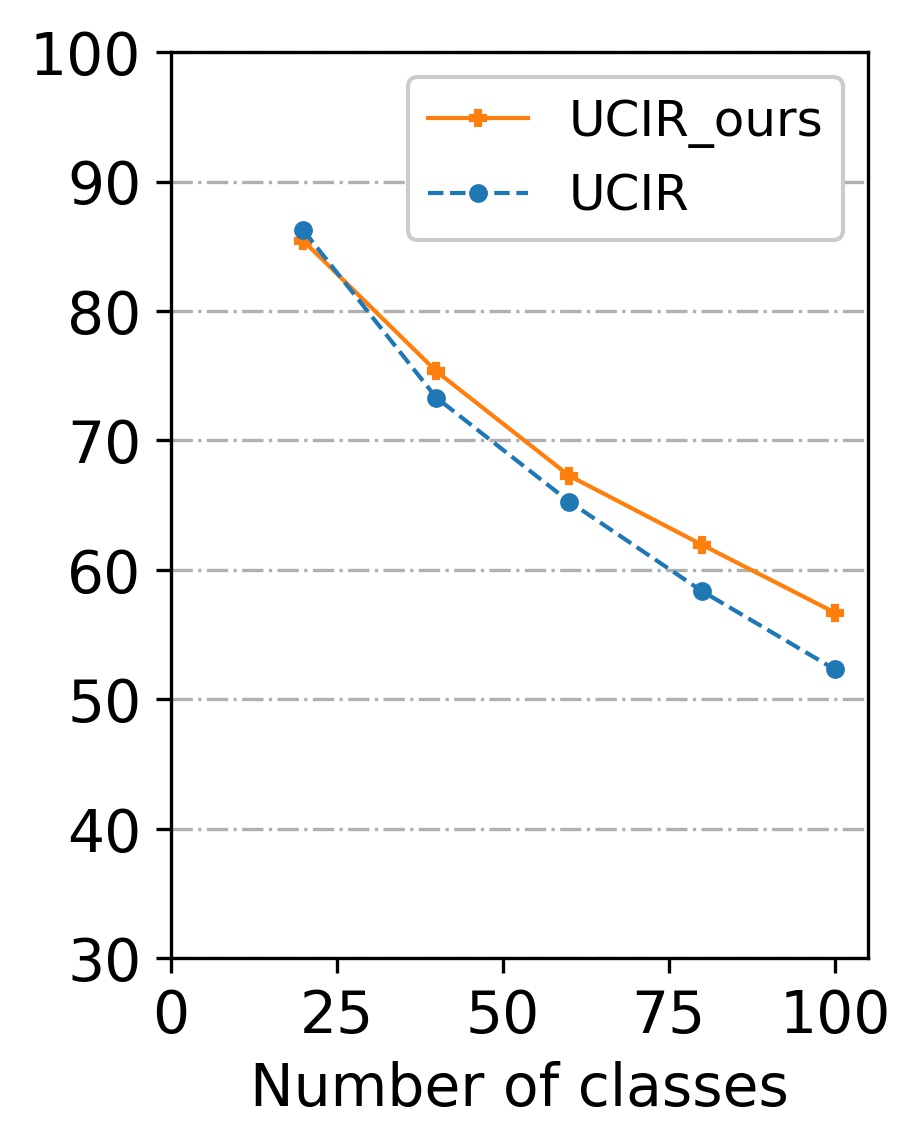}
    \includegraphics[width=0.24\linewidth]{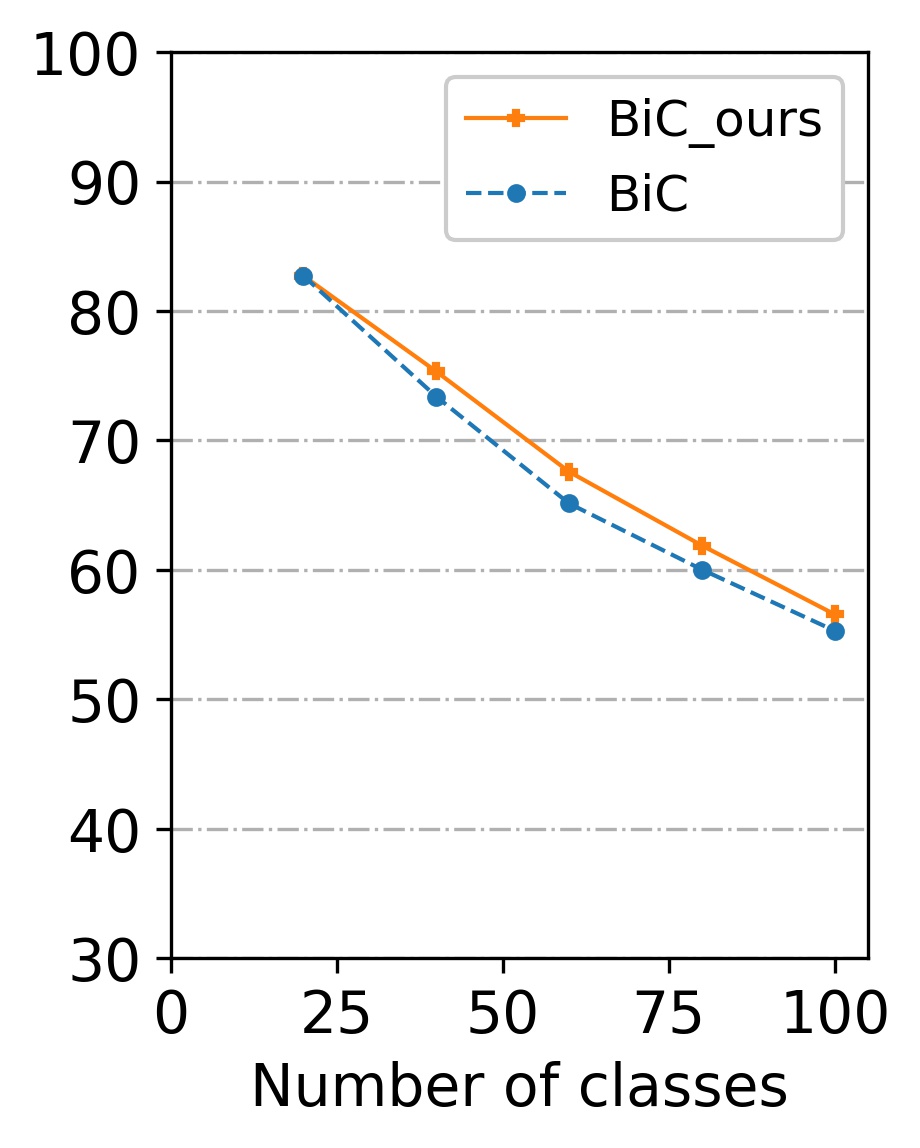}
    \caption{Continual learning of 5 (first row), 10 (second row), and 20 (third row) new classes at each round with CIFAR100 dataset. Columns 1-4 (blue curve): performance of LwF, iCaRL, UCIR, BiC; Columns 1-4 (orange curve): performance of the proposed method built on the corresponding baseline.}
    \label{fig:cifar100}
\end{figure}
\begin{figure}[!btp]
    \centering
    \includegraphics[width=0.24\linewidth, height=0.3\linewidth]{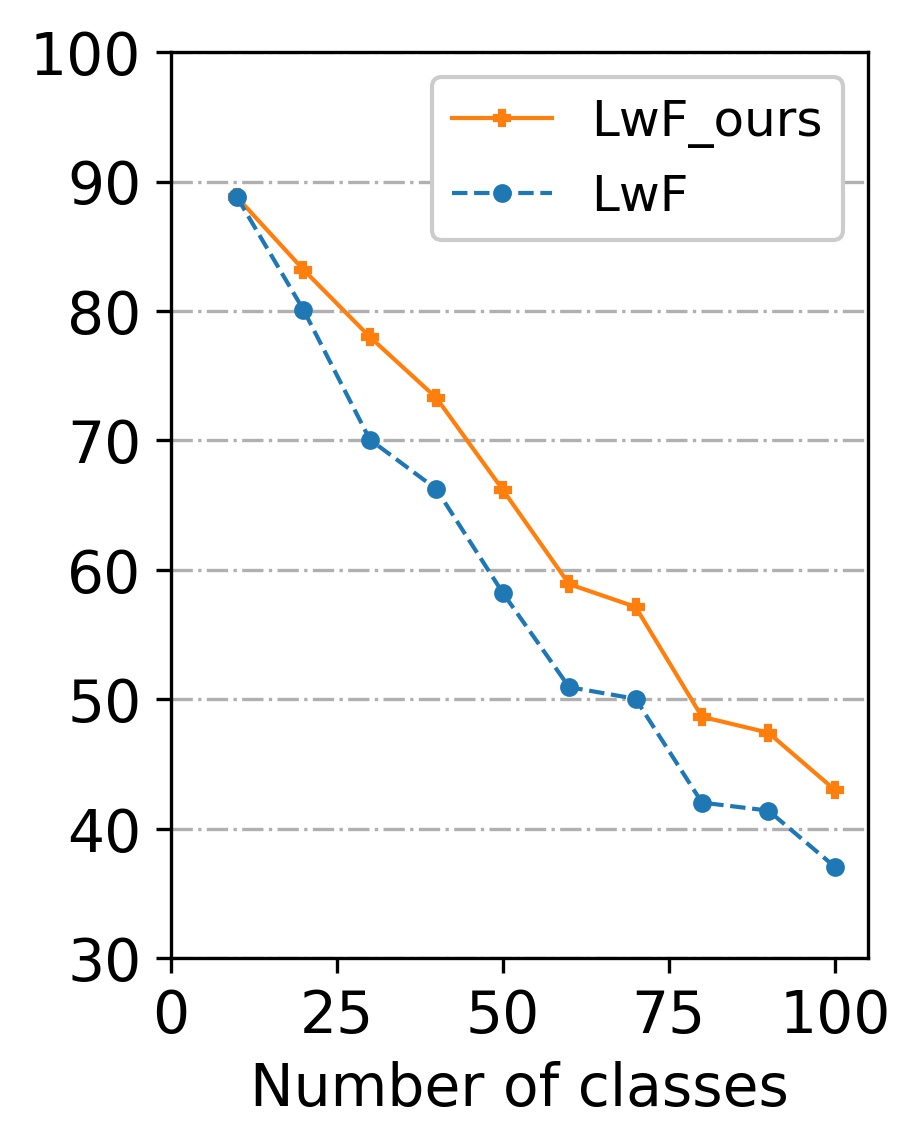}
    \includegraphics[width=0.24\linewidth, height=0.3\linewidth]{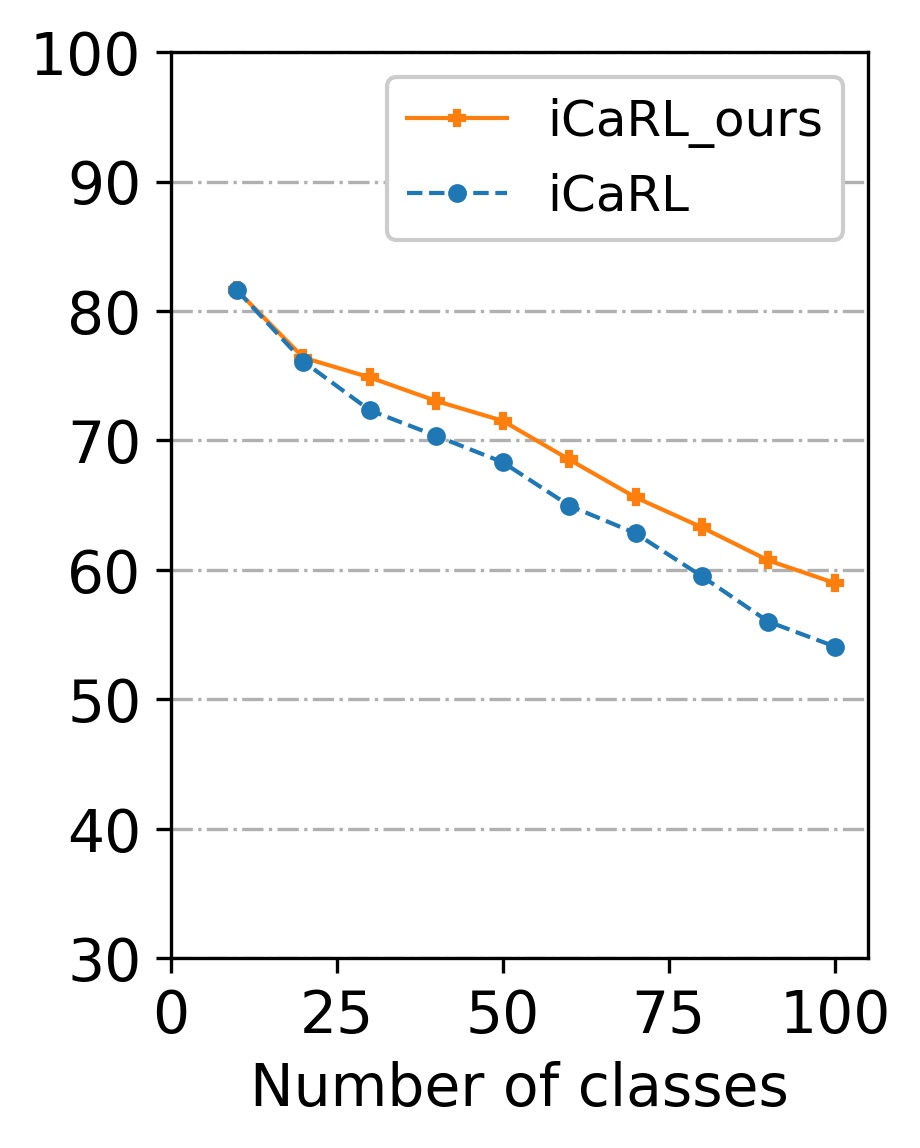}
    \includegraphics[width=0.24\linewidth, height=0.3\linewidth]{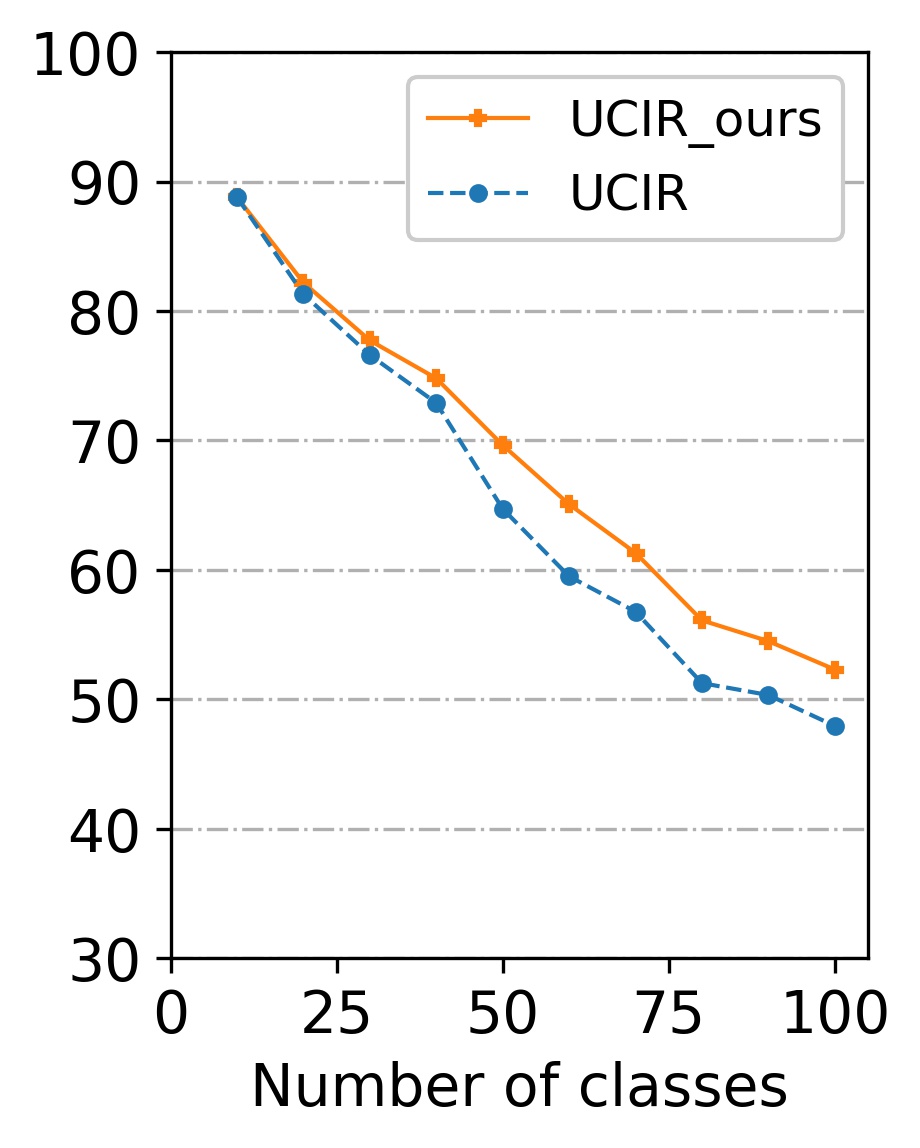}
    \includegraphics[width=0.24\linewidth, height=0.3\linewidth]{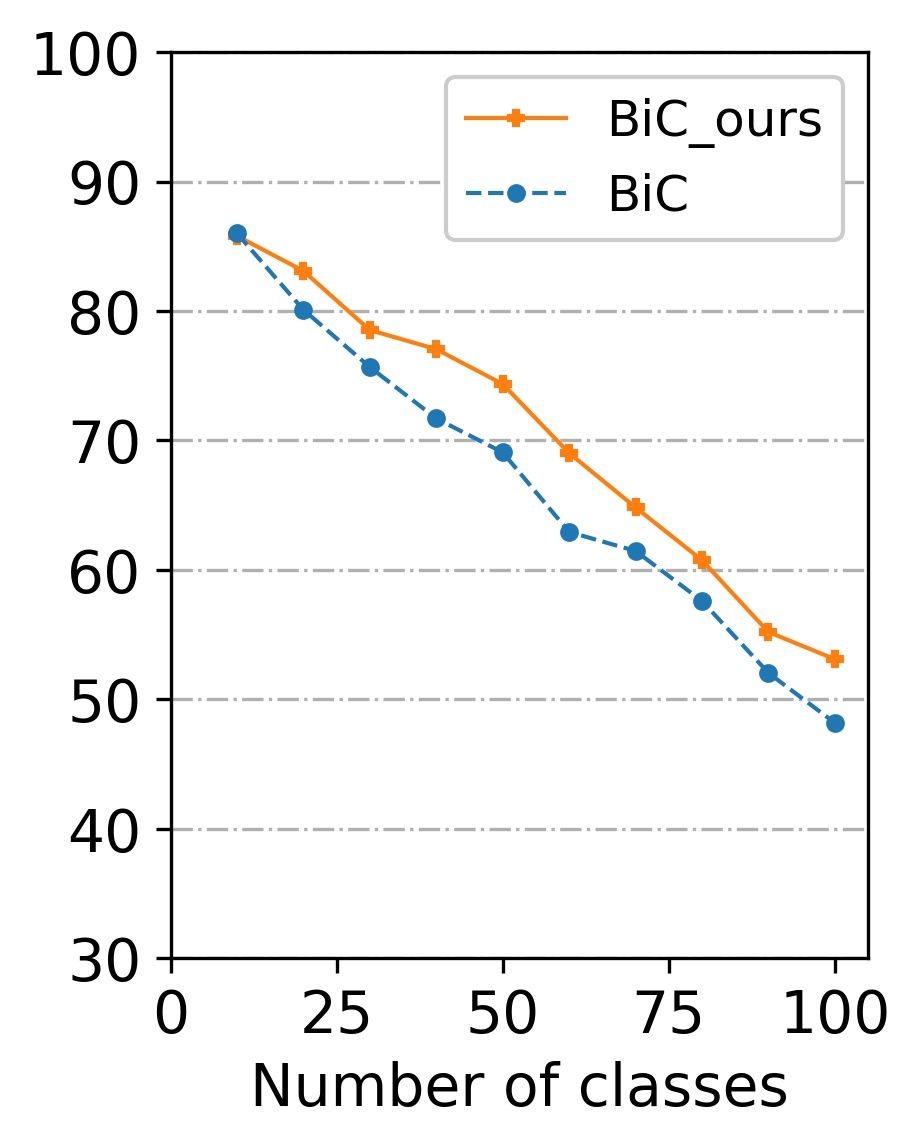}
    \caption{Continual learning of 10 new classes at each round with mini-ImageNet dataset. }
    \label{fig:mini_imagenet}
\end{figure}
\begin{figure}[!btp]
    \centering
    \includegraphics[width=0.32\linewidth]{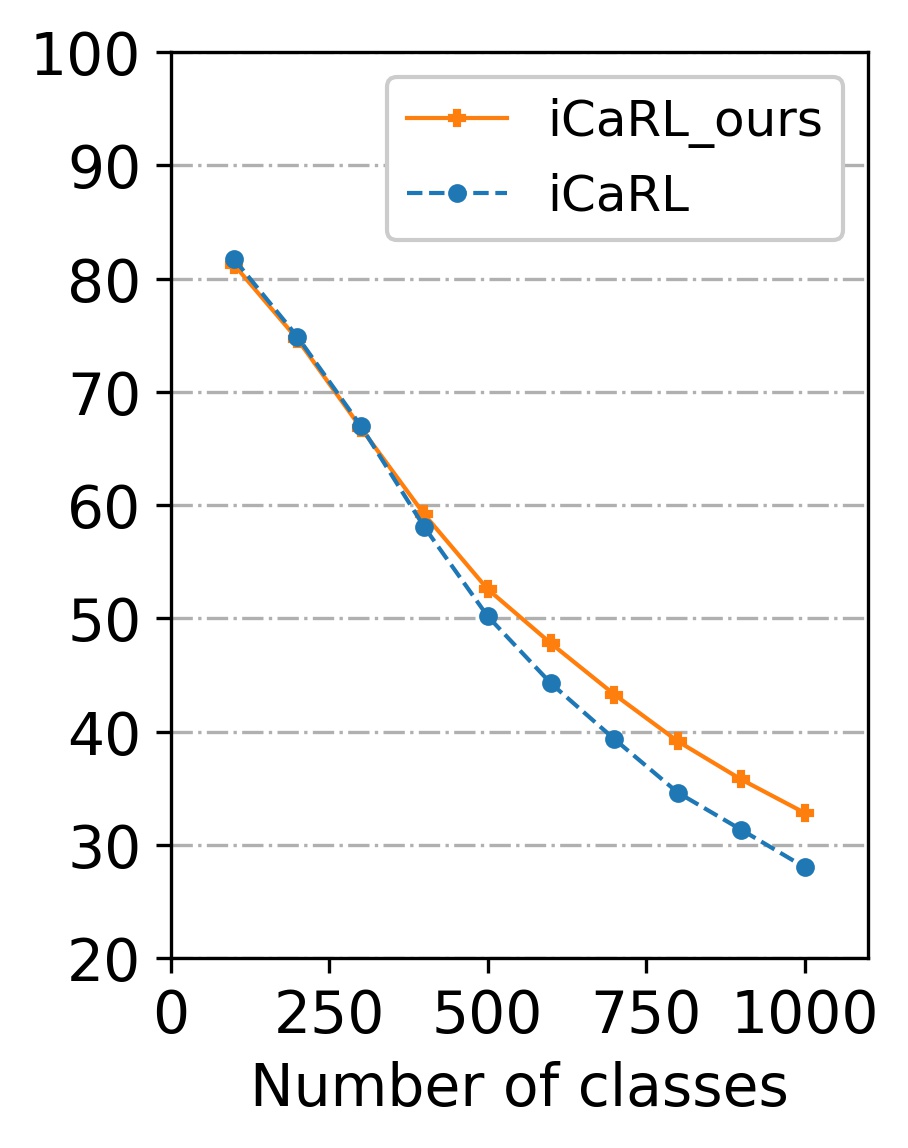}
    \includegraphics[width=0.32\linewidth]{imagenet_subset_top1_UCIR_ours.jpg}
    \includegraphics[width=0.32\linewidth]{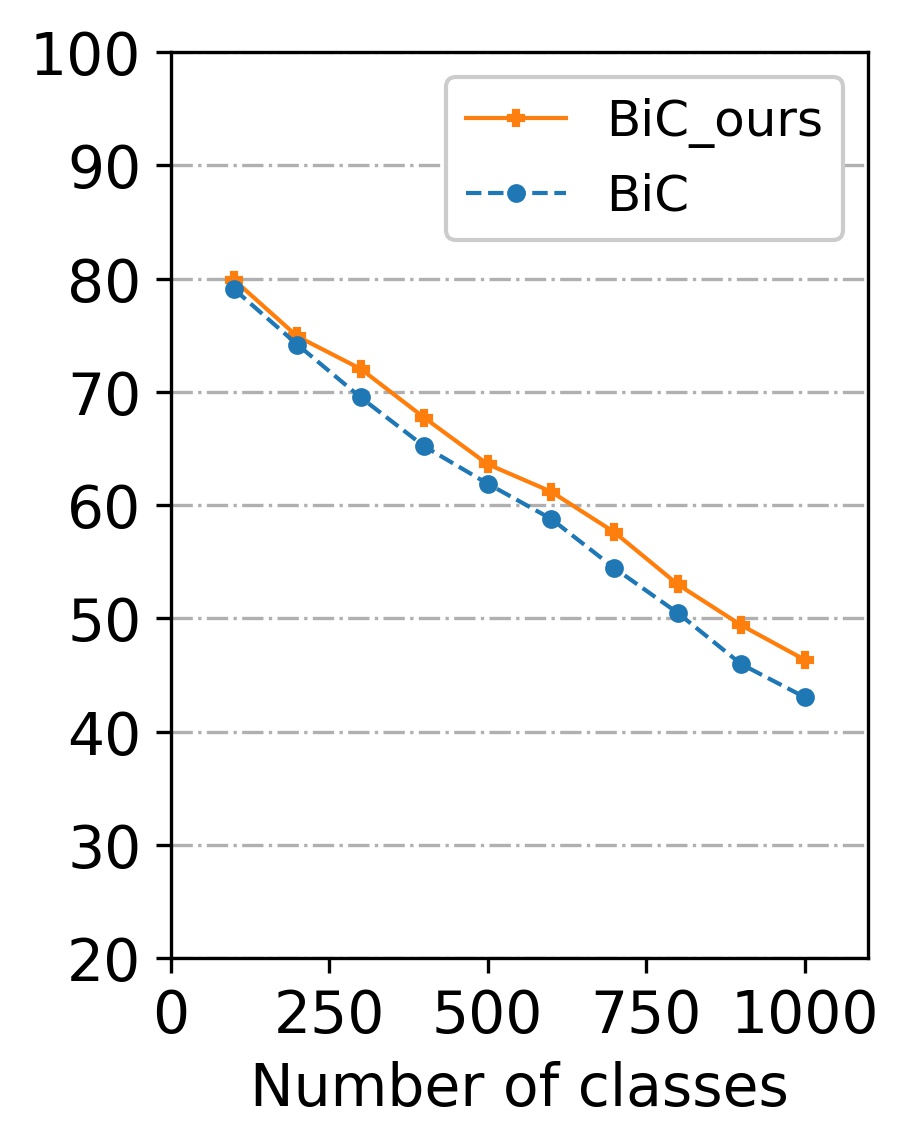}
    \caption{Continual learning of 100 new classes at each round with ImageNet dataset. }
    \label{fig:imagenet}
\end{figure}

The proposed discriminative distillation can be plugged into most continual learning methods. Therefore, the effectiveness of the proposed distillation is evaluated by combining it respectively with existing continual learning methods, including LwF~\cite{li2017learning}, iCaRL~\cite{rebuffi2017icarl}, UCIR~\cite{hou2019learning}, and BiC~\cite{wu2019large}. All the four methods are distillation-based, and therefore the only difference between each baseline and the corresponding proposed method is the inclusion of the discriminative distillation loss term during classifier training. 
The inference method proposed in the original papers were adopted during testing (nearest-mean-of-exemplars for iCaRL, and softmax output for LwF, UCIR, and BiC).
The evaluation was firstly performed on the CIFAR100 dataset. As shown in Figure~\ref{fig:cifar100}, when continually learning 5 classes (first row), 10 classes (second row), and 20 classes (third row) at each round respectively, each baseline method was clearly outperformed by the combination of the proposed discriminative distillation with the baseline, with around absolute $2\% \sim 5\%$ better in accuracy at each round of continual learning. The consistent improvement in classification performance built on different continual learning methods supports the effectiveness of the proposed discriminative distillation for continual learning.  
Similar results were obtained from experiments on the mini-ImageNet (Figure~\ref{fig:mini_imagenet}) and the ImageNet dataset (Figure~\ref{fig:imagenet}), suggesting that the proposed discriminative distillation is effective in various continual learning tasks with different scales of new classes at each learning round.

To further investigate the effect of the proposed discriminative distillation on reducing the confusions between similar classes during classification, the total reduction in classification error compared to the baseline method at each round of learning was divided into two parts, one relevant to class confusion and the other relevant to catastrophic forgetting. For CIFAR100, it is well-known that the dataset contains 20 meta-classes (e.g., human, flowers, vehicles, etc.) and each meta-class contains 5 similar classes (e.g., baby, boy, girl, man, and woman). At any round of continual learning, if the trained new classifier mis-classify one test image into another class which shares the same meta-class, such an classification error is considered partly due to class confusion (Table~\ref{tab:confusion}, `Confusion'). Otherwise, if one test image of any old class is mis-classified to another class belonging to a different meta-class, this classification error is considered partly due to catastrophic forgetting (Table~\ref{tab:confusion}, `Forgetting'). From Table~\ref{tab:confusion} (second row), it can be observed that the proposed discriminative distillation did help reduce the class confusion error compared to the corresponding baseline (first row) at various learning rounds. 
In addition, the discriminative distillation can also help reduce the catastrophic forgetting  (Table~\ref{tab:confusion}, last two rows), consistent with the previously reported result based on the distillation of only new classes from the expert classifier~\cite{li2020continual}. 

\begin{figure*}[t]
    \centering
    \includegraphics[width=0.48\linewidth]{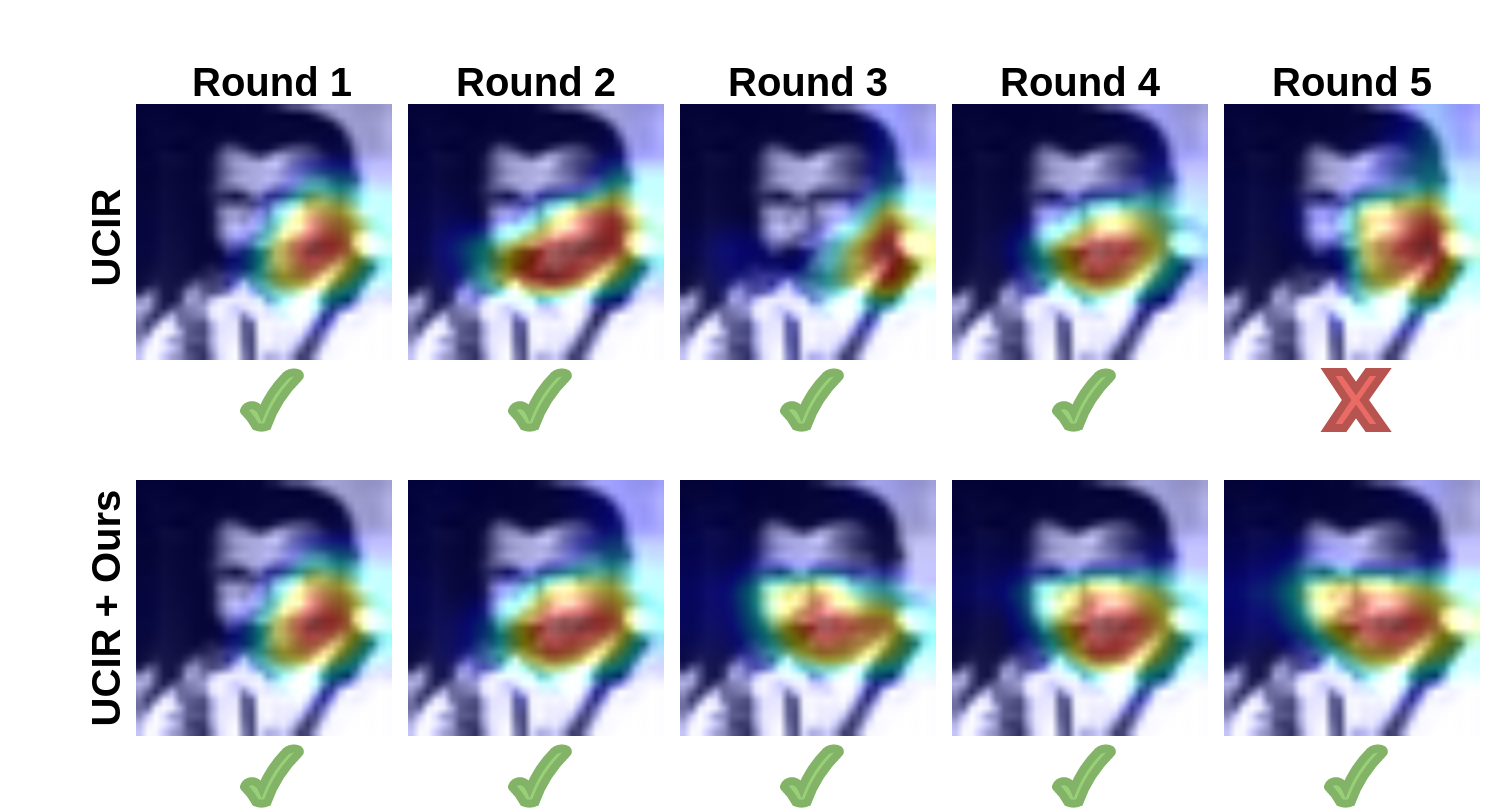}
    \includegraphics[width=0.48\linewidth]{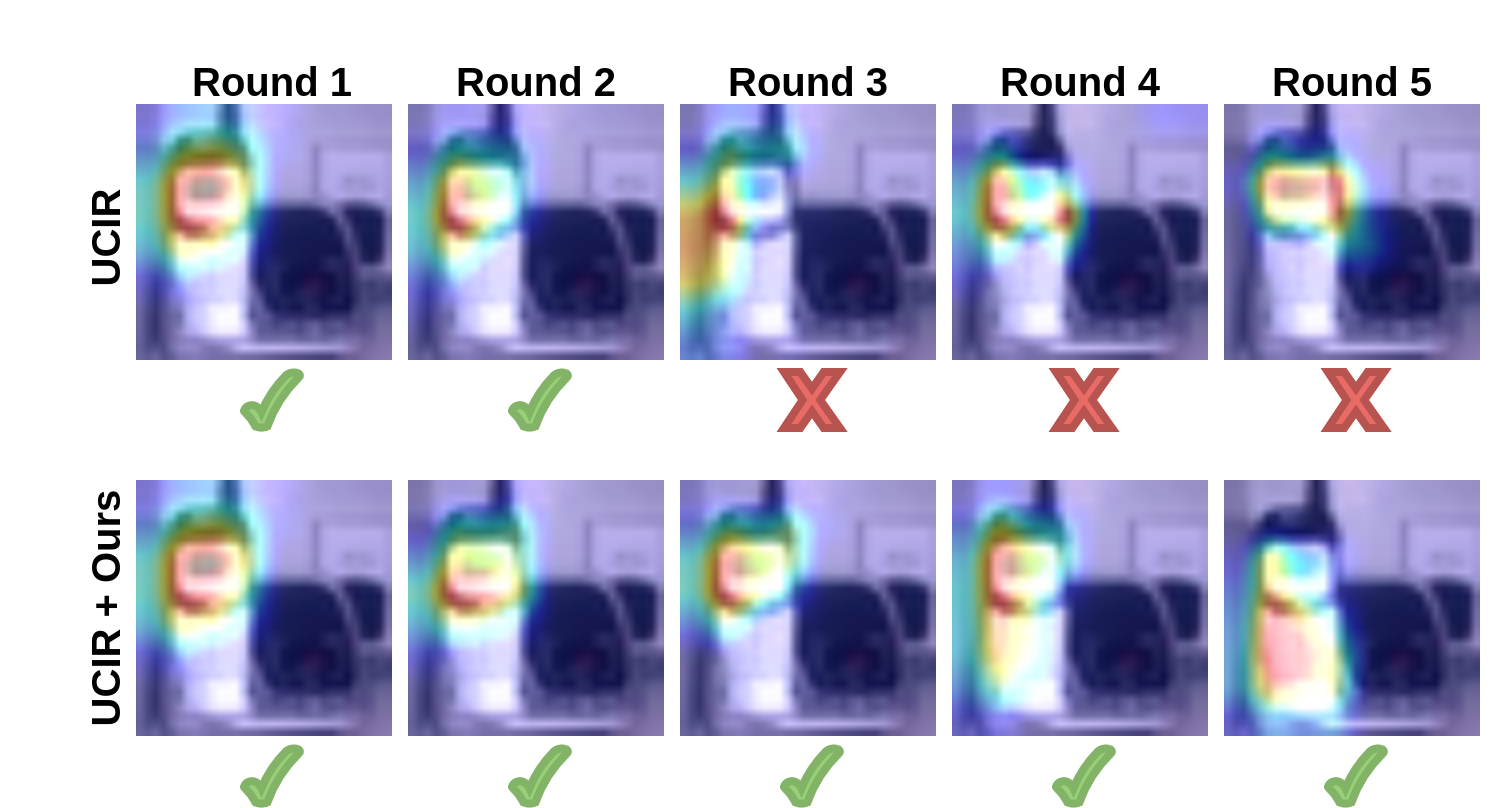}
    \vspace{0.2cm}
    \includegraphics[width=0.47\linewidth]{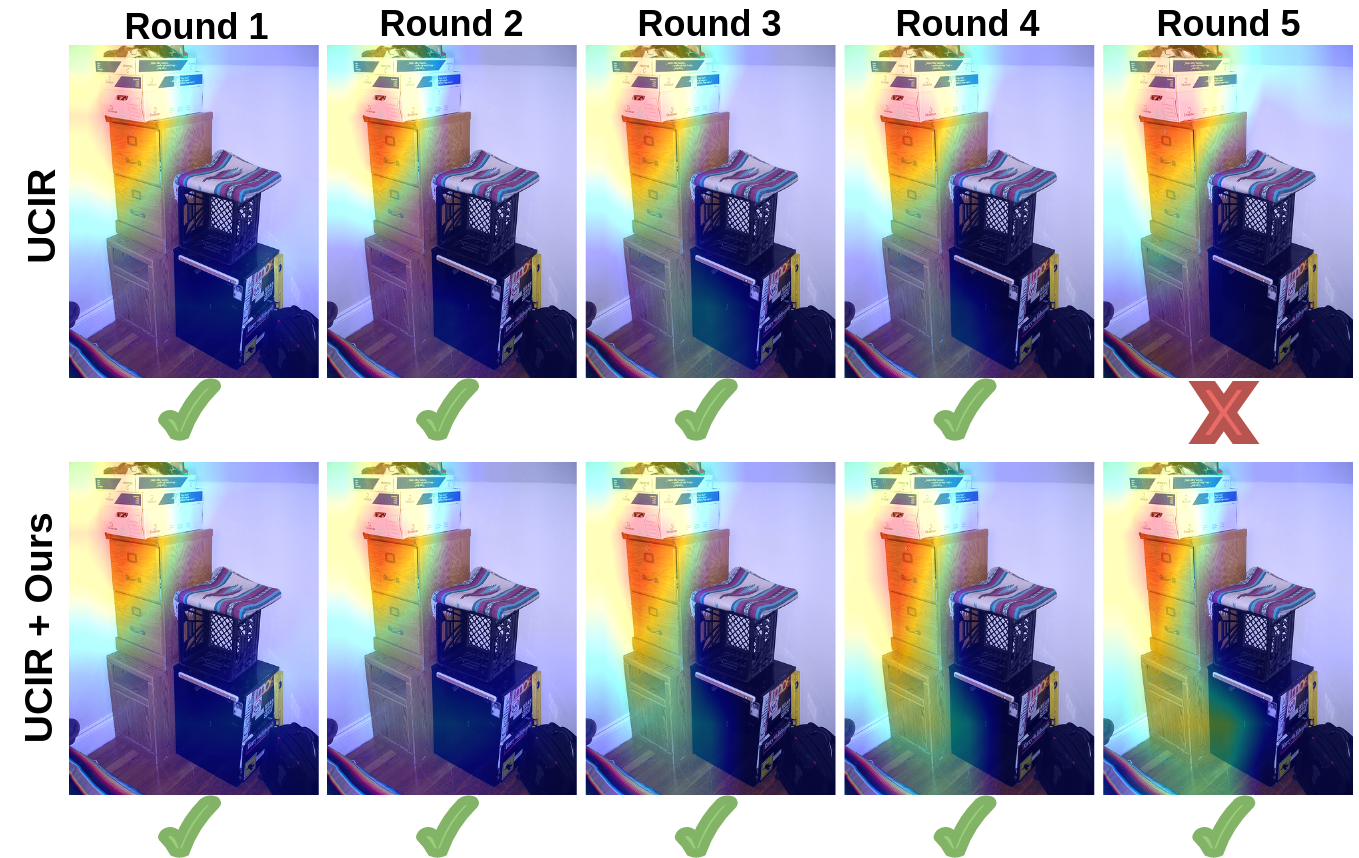}
    \vspace{0.2cm}
    \includegraphics[width=0.46\linewidth]{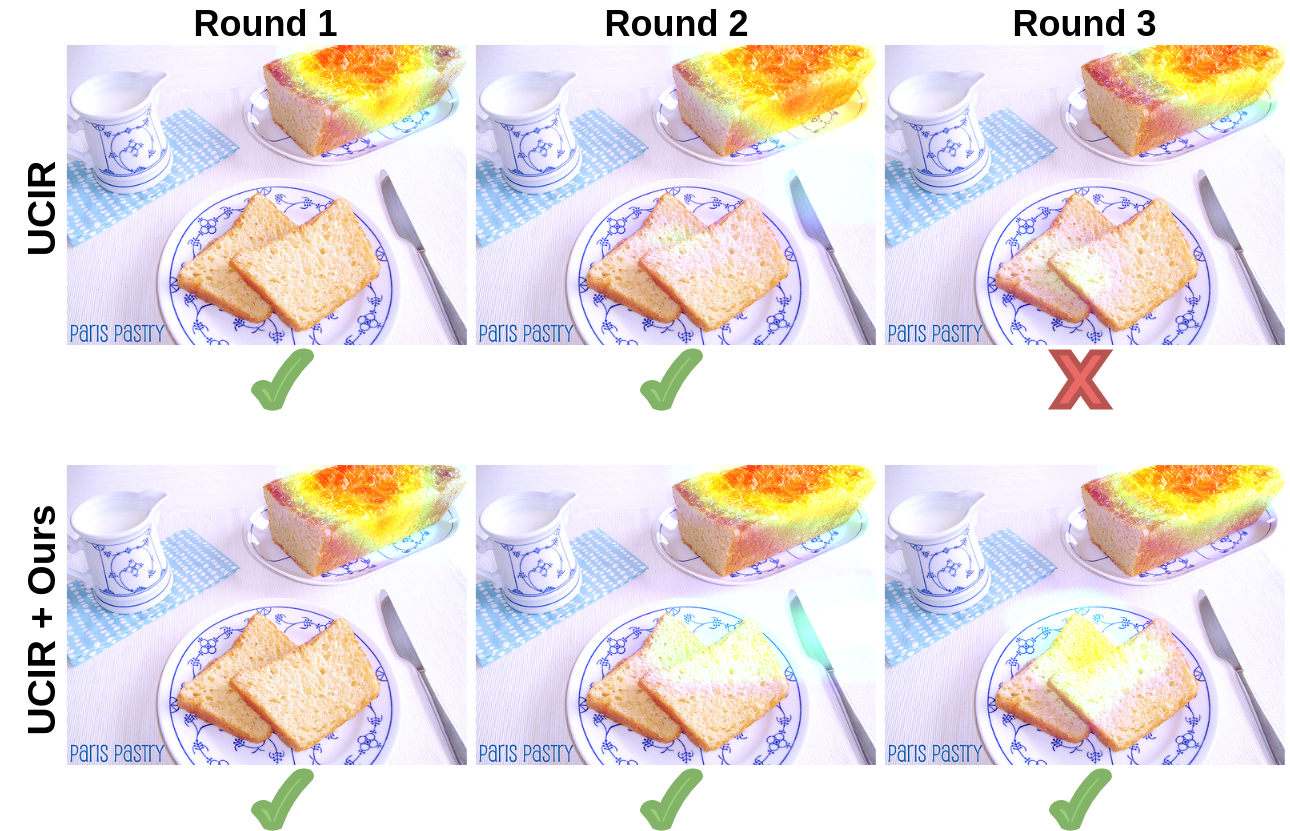}
   \caption{The demonstrative attention maps over learning rounds from the baseline UCIR (upper row) and the correspondingly proposed method (bottom row). The input images are from CIFAR100 and mini-ImageNet, and each attention map (heatmap overlapped on the input image) for the ground-truth class was generated by the Grad-CAM method~\cite{selvaraju2017grad} from the trained classifier at each learning round. The tick or cross under each image represents the classification result. 
   These examples show that keeping old knowledge does not guarantee the correct classification results. The proposed method helps the model learn to focus on the more discriminative image regions.}
   \label{fig:cam}
\end{figure*}

The effect of the discriminative distillation was also visually confirmed with demonstrative examples of attention map changes over learning rounds (Figure~\ref{fig:cam}). For example, while the classifier trained based on the baseline UCIR can  attend to some part of the `man' face over learning rounds, this test image was mis-classified at the last round (Figure~\ref{fig:cam}, top row, left half). This suggests that the mis-classification is probably not due to forgetting of the old knowledge (otherwise the attended region at last learning round would be much different from that at the first round). In comparison, the classifier based on the correspondingly proposed method learned to attend to larger face regions and can correctly classify the image over all rounds (Figure~\ref{fig:cam}, second row, left half), probably because the expert classifier learned to find that more face regions are necessary in order to discriminate different types of human faces (e.g., `man' vs. `women') and such discriminative knowledge was distilled from the expert classifier to the new classifier during continual learning. Similar results can be obtained from the other three examples (Figure~\ref{fig:cam}, `phone' image on top right, `file cabinet' image on bottom left, and `bread' image on bottom right).


\begin{table}
    \centering
    \caption{Effect of the proposed distillation on the reduction of class confusion and catastrophic forgetting. Each value is the number of images incorrectly classified by the new classifier at that learning round.}
    \begin{tabular}{@{}clcccc@{}}
    \toprule
        \multirow{2}{*}{Types} & \multirow{2}{*}{Methods} & \multicolumn{4}{c}{Learning rounds}\\
        \cmidrule{3-6} 
        && 2 & 3 & 4 & 5 \\
        \midrule
        \multirow{2}{*}{Confusion}  &UCIR &193 &547 &1037 &1525 \\ 
                &UCIR+ours &178 &524 &986 &1399 \\
        \hline
        \multirow{2}{*}{Forgetting}  &UCIR &328 &1002 &1813 &2706 \\ 
                &UCIR+ours &309 &944 &1633 &2484 \\ 
    \bottomrule
    \end{tabular}
\label{tab:confusion}
\end{table}



\subsection{Ablation study}

\begin{figure}[]
\begin{center}
   \includegraphics[width=0.45\linewidth]{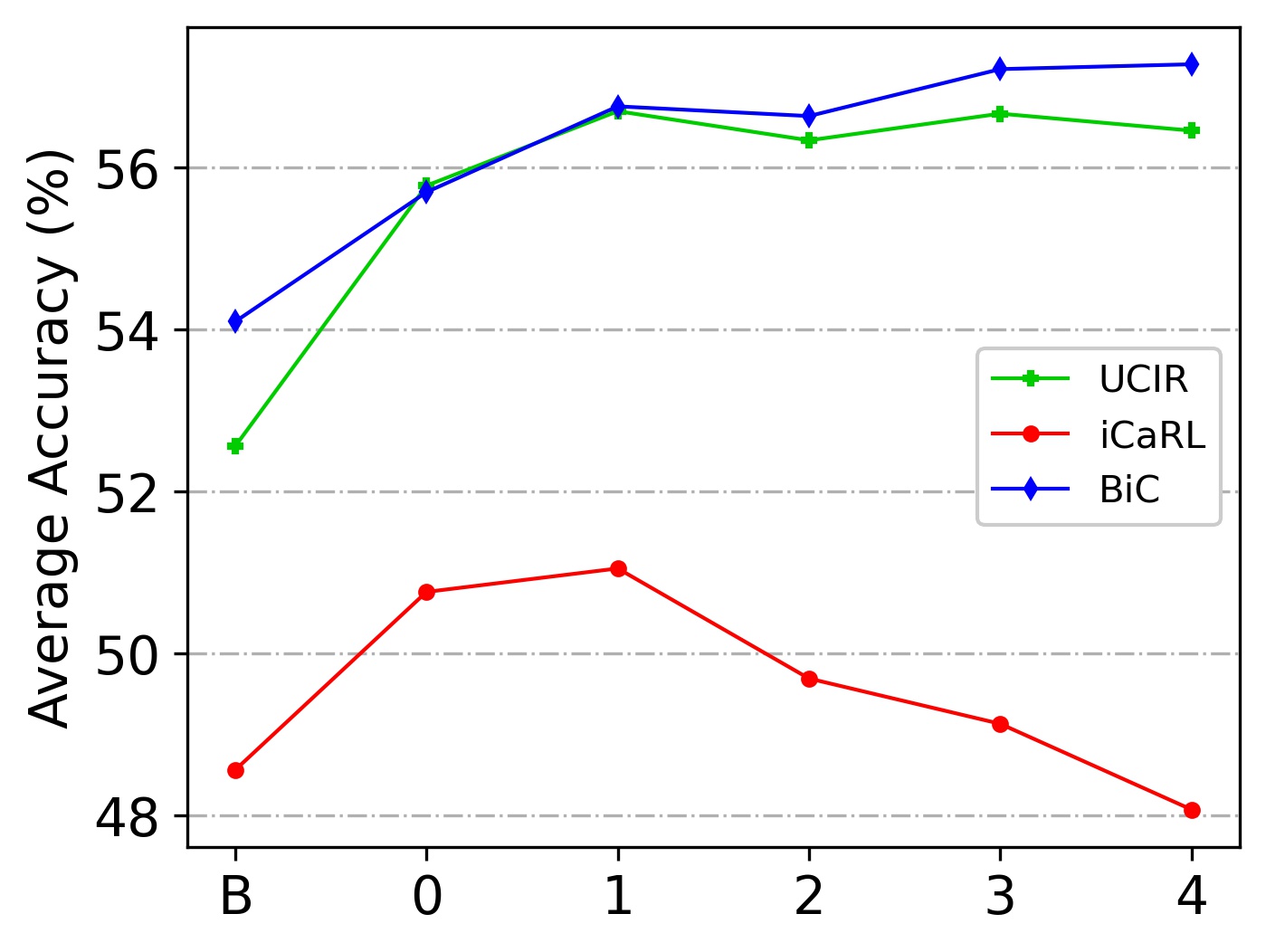}
   \includegraphics[width=0.45\linewidth]{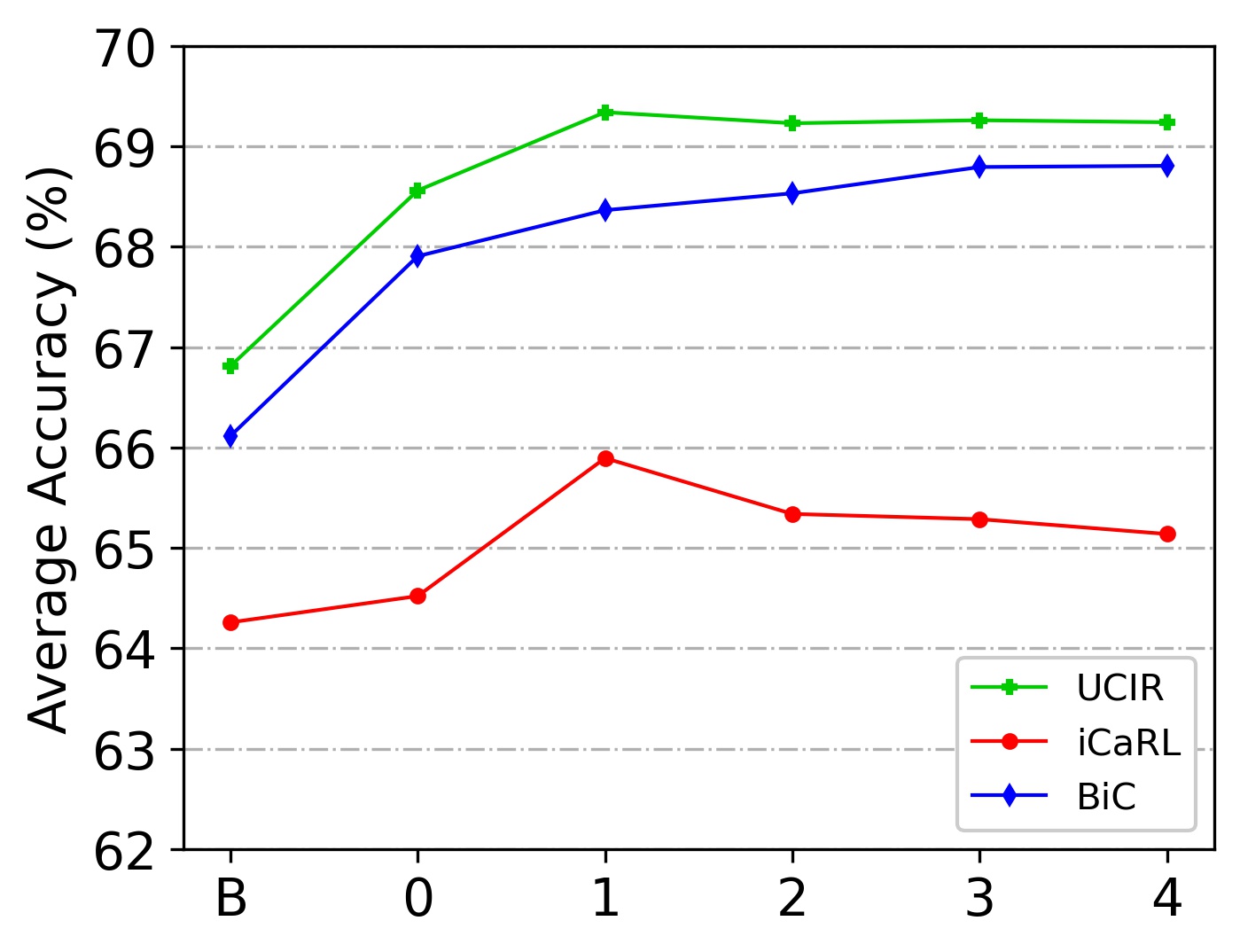}
\end{center}
   \caption{Ablation study built on different baseline methods, with 20 new classes learned at each round on the CIFAR100 dataset. X-axis: number of  similar old classes selected for each new class during continual learning; `0' means the expert classifier only learns new classes, and 
   `B' (short for `Baseline') means the expert classifier is not applied during learning. Y-axis: the mean classification accuracy over all the classes at the final round (Left), or the average of mean class accuracy over all learning rounds (Right).  
   }
\label{fig:ablation}
\end{figure}

The effect of the discriminative distillation is further evaluated with a series of ablation study built on different baseline methods. As Figure~\ref{fig:ablation} shows, compared to the baselines (`Baseline' on the X-axis) and the dual distillation which does not learn any old classes in the expert classifier (`0' on the X-axis), learning to classify both new and similar old classes by the expert classifier and then distilling the discriminative knowledge to the new classifier (`1' to `4' on X-axis) often improves the continual learning performance, either at the final learning round (Figure~\ref{fig:ablation}, Left) or over all the learning rounds (Figure~\ref{fig:ablation}, Right). Adding more old classes for the expert classifier does not always improve the performance of the new classifier (Figure~\ref{fig:ablation}, Left, red curve), maybe because the inclusion of more old classes distracts the expert classifier from learning the most discriminative features between confusing classes.


\section{Conclusions}
Continual learning may be affected not only by catastrophic forgetting of old knowledge, but also by the class confusion between old and new knowledge. This study proposes a simple but effective discriminative distillation strategy to help the classifier handle both issues during continual learning. The distillation component can be flexibly embedded into existing approaches to continual learning. Initial experiments on natural image classification datasets shows that explicitly handling the class confusion issue can further improve continual learning performance. This suggests that both catastrophic forgetting and class confusion may need to be considered in future study of continual learning.

\bibliographystyle{ACM-Reference-Format}
\bibliography{mybib}

\appendix









\end{document}